%% file: ex_article_post_feedback.tex
\documentclass[review,hidelinks,onefignum,onetabnum]{siamart220329}

\input{ex_shared}

\ifpdf
\hypersetup{
  pdftitle={Common pitfalls to avoid while using multiobjective optimization in machine learning},
  pdfauthor={J. Akhter, P.D. Fährmann, K. Sonntag, S. Peitz, D. Schwietert}
}
\fi

\usepackage{todonotes}
\newcommand{\revisioncolor}{black}
\usepackage[normalem]{ulem}

\begin{document}

\nolinenumbers
\maketitle
\input{abstract}                                          
\section{Introduction}
\label{sec:introduction}
\input{chapters_post_feedback/introduction1}                              
\section{Multiobjective optimization}
\label{sec:MOO}
\input{chapters_post_feedback/theory/MOO}                                
\section{Multiobjective optimization and deep learning}
\label{sec:MOO&Dl}
\input{chapters_post_feedback/theory/MOOFDL}                            
\section{Pitfalls}
\label{sec:results}
\input{chapters_post_feedback/Results/pitfalls}

\section{Conclusion}
\label{sec:conclusion}
\input{chapters_post_feedback/conclusion}



\section*{Acknowledgments}
This research was conducted within the AI junior research group ``Multicriteria Machine Learning'' which is funded by the German Federal Ministry of Education and Research (BMBF).


\end{document}

%% file: ex_shared.tex
\usepackage{hyperref}       
\usepackage{url}            
\usepackage{booktabs}       
\usepackage{bm}
\usepackage{amssymb}
\usepackage{amsfonts}  
\usepackage{xcolor}    
\definecolor{revisioncolor}{rgb}{0,0,0} 
\renewcommand{\textcolor}[2]{#2}        
\usepackage{nicefrac}       
\usepackage{microtype}      
\usepackage{lipsum}		
\usepackage{graphicx}
\usepackage{subcaption}
\usepackage{doi}
\usepackage{enumerate} 
\usepackage{tikz}
\usepackage{enumitem}
\usepackage{caption}
\usepackage{comment}
\usepackage{mathtools}
\usepackage{thmtools}
\usepackage{float}

\newcommand{\R}{\mathbb{R}}

\newcommand{\Loss}{\mathcal{L}}
\DeclareMathOperator{\argmin}{argmin}
\DeclareMathAlphabet\mathbfcal{OMS}{cmsy}{b}{n}

\ifpdf
  \DeclareGraphicsExtensions{.eps,.pdf,.png,.jpg}
\else
  \DeclareGraphicsExtensions{.eps}
\fi

\newsiamthm{mythm}{Theorem}
\newsiamthm{myprop}{Proposition}
\newsiamthm{mydef}{Definition}
\newsiamthm{mycorollary}{Corollary}
\newsiamthm{mylemma}{Lemma}
\newsiamthm{myremark}{Remark}
\newsiamthm{myassump}{Assumption}
\newsiamthm{myexample}{Example}

\headers{Pitfalls when using MOO in ML}{J. Akhter, P.D. F\"ahrmann, K. Sonntag, S. Peitz, D. Schwietert}

\title{Common pitfalls to avoid while using multiobjective optimization in machine learning\thanks{Submitted to the editors DATE.
\funding{This work was funded by the German Federal Ministry of Education and Research (BMBF) within the AI junior research group ``Multicriteria Machine Learning''.}}}

\author{Junaid Akhter\thanks{Department of Computer Science, TU Dortmund, Germany (\email{junaid.akhter@tu-dortmund.de}, \email{sebastian.peitz@tu-dortmund.de}, \email{paul.faehrmann@tu-dortmund.de}).}
\and Paul David F\"ahrmann \footnotemark [2]
\and Konstantin Sonntag\thanks{Department of Computer Science, University of Paderborn, Germany (\email{konstantin.sonnntag@uni-paderborn.de}, \email{danielki@mail.uni-paderborn.de})}
\and Sebastian Peitz\footnotemark[2]
\and Daniel Schwietert\footnotemark[3]
}

\usepackage{amsopn}


%% file: abstract.tex
\begin{abstract}
Recently, there has been an increasing interest in the application of multiobjective optimization (MOO) in machine learning (ML). This interest is driven by the numerous real-life situations where multiple objectives must be optimized simultaneously. A key aspect of MOO is the existence of a Pareto set, rather than a single optimal solution, which represents the optimal trade-offs between different objectives. Despite its potential, there is a noticeable lack of satisfactory literature serving as an entry-level guide for ML practitioners aiming to apply MOO effectively. In this paper, our goal is to provide such a resource and highlight pitfalls to avoid. We begin by establishing the groundwork for MOO, focusing on well-known approaches such as the weighted sum (WS) method, alongside more advanced techniques like the multiobjective gradient descent algorithm (MGDA). We critically review existing studies across various ML fields where MOO has been applied and identify challenges that can lead to incorrect interpretations. One of these fields is physics informed neural networks (PINNs), which we use as a guiding example to carefully construct experiments illustrating these pitfalls. By comparing WS and MGDA with one of the most common evolutionary algorithms, NSGA-II, we demonstrate that difficulties can arise regardless of the specific MOO method used. We emphasize the importance of understanding the specific problem, the objective space, and the selected MOO method, while also noting that neglecting factors such as convergence criteria can result in misleading experiments.
\end{abstract}

%% file: chapters_post_feedback/introduction1.tex
Over the past decade, machine learning (ML) has seen remarkable growth and widespread adoption, both in research and applications. This surge is partly attributed to artificial neural networks (ANNs) because of their ability to model any continuous function \cite{universal_csanad_csaji_approximation_2001, universal_2_DBLP:journals/corr/abs-1709-02540}, given a sufficiently large network architecture. Another major catalyst for this growth owes to the availability of powerful computational resources.

By adding multiple layers to ANNs, we acquire a deep neural network (DNN) and hence transition into \textit{deep learning} (DL), enabling the extraction of complex patterns from data. Generally speaking, the task in ML/DL is to train a model using a training data set, such that it is capable of making accurate predictions and generalizes to unseen data. This task translates to optimizing a performance metric, also referred to as \textit{loss function} or an \textit{objective}. There are numerous efficient single objective optimization (SOO) techniques \cite{ML_algorithms_2021, math11112466} for DNN training, the ADAM \cite{kingma2017adam} algorithm being one of the most popular choices.

In classical optimization, the consideration of multiple, conflicting criteria is a long-known challenge. There are many examples where \textit{multiobjective optimization} (MOO) is very important, for instance in production processes, where one wants to maximize the quality of a product while minimizing its cost. Hence, instead of a single optimal solution, there exists a set of optimal trade-offs called \textit{Pareto set} \cite{Miettinen1998}. The Pareto set represents the set of points for which no other feasible solutions exist that can improve one objective without degrading at least one of the other objectives. Areas where MOO plays a huge role include finance, economics, health systems, logistics, agriculture, and applied engineering. 

\textcolor{\revisioncolor}{Bringing MOO techniques and ML together is a natural research direction, as conflicting criteria occur often, such as the need to minimize loss, incorporating domain knowledge, reducing architectural size, or tackling distinct tasks in multi-task learning. Recently, this field has gained momentum, supported by multiple studies \cite{On_the_pareto_front, Bieker_Katharina_Peitz, hotegni2024multiobjective, sener_multi-task_nodate, amakor2024multiobjective}. In the study \cite{sener_multi-task_nodate}, a multiobjective gradient descent algorithm (MGDA) \cite{desideri2012mgda, desideri2014pareto} was applied to multi-task optimization in the context of digit classification. Our investigations however reveal that the sophisticated architecture of their model effectively mitigates the task conflicts, allowing for single solutions optimal for multiple tasks.}

\textcolor{\revisioncolor}{Learning solutions to problems governed by physical laws with ML has led to the development of physics-informed neural networks (PINNs) \cite{karniadakis_physics-informed_2021, 712178, raissi2017physicsinformeddeeplearning, blechschmidt2021three, cuomo2022scientific}. PINNs allow for a natural application of MOO methods by integrating domain knowledge with measurement data. This integration is typically achieved by defining two loss functions: the \textit{physics loss}, derived from governing equations (such as ordinary and partial differential equations), and the \textit{data loss}, based on traditional supervised learning. In addition to incorporating domain-specific knowledge, PINNs can also embed fundamental physical principles, such as conservation laws \cite{JAGTAP2020113028, wu2021modified, baez2024guaranteeing, WU2022112143, LIN2022111053, liu2024discontinuity} and Hamiltonian dynamics \cite{MORADI20235152, greydanus2019hamiltonianneuralnetworks}. These principles further constrain the solution space to ensure physically realistic outcomes.
Despite the huge success in solving various systems like multi-physics systems \cite{coulaud:hal-04225990}, fluid dynamics \cite{Cai2021PINNsReview}, control problems \cite{antonelo2024physics}, modeling infection \cite{berkhahn2022covid}, fractional PDEs \cite{fPINNS}, integro differential equations \cite{integro_diff_pinns}, parametrized PDEs \cite{parametric_pdepinns}, or stochastic PDEs \cite{Stochastic_pdepinns}, PINNs face several issues like hyper-parameter tuning \cite{bishof_2021} and slow convergence.}

\textcolor{\revisioncolor}{Several studies, such as \cite{bishof_2021, On_the_pareto_front, HELDMANN2023112211, s23218665}, have applied MOO with PINNs with claims of performance improvement. Each of these works uses a form of scalarization, primarily through the weighted sum (WS) method, to handle multiple objectives. While \cite{bishof_2021, HELDMANN2023112211, s23218665} apply WS without exploring the full Pareto front, \cite{On_the_pareto_front} goes further by systematically investigating the Pareto front in the context of PINNs. However, we observed potential gaps in the interpretation of these Pareto fronts and the methodology used to obtain them. We reveal that some conclusions drawn in these studies may be inaccurate, as their methodologies may not fully align with MOO principles. Misinterpretations and misapplications of MOO can lead to suboptimal decision-making and affect the validity of the identified trade-offs. This highlights the need for a better understanding of MOO principles in DL. The goal of this paper is to present guidelines for the correct application of MOO and to point out pitfalls one should avoid in interpreting results and training DL models, particularly PINNs.}

\subsection{Current literature in the area of Multiobjective Deep Learning}
The landscape of MOO in DL is marked by significant contributions across various methodologies and applications \textcolor{\revisioncolor}{ \cite{On_the_pareto_front, Bieker_Katharina_Peitz, ma2020efficient, hotegni2024multiobjective, lopezruiz2024multiobjective, sener_multi-task_nodate, amakor2024multiobjective, DBLP:journals/corr/abs-2001-06782}}. A notable paper in the context of multi-task learning (MTL) is \cite{sener_multi-task_nodate}, where one neural network is trained to solve two learning tasks using MGDA. A MOO algorithm for DNNs, employing a modified weighted Chebyshev scalarization to simplify the optimization of multiple tasks into sequential single-objective problems, is proposed in \cite{hotegni2024multiobjective}. A model-agnostic method named Adamize \cite{mitrevski2021momentumbased} adapts the benefits of the Adam optimizer to multiobjective problems. Additionally, efforts have been made to develop approaches that generate the entire Pareto front in a single run \cite{ruchte2021scalable}. For an in-depth analysis of MTL in computer vision, one can look into \cite{Vandenhende_2021}. {\color{\revisioncolor} In the area of PINNs, gradient-based MOO approaches have proven particularly useful in the balancing act between different losses \cite{On_the_pareto_front, bishof_2021, LIU2021112, McClenny_2023, XIANG202211, basir2023adaptiveaugmentedlagrangianmethod, SON2023126424, ANAGNOSTOPOULOS2024116805, Maddu_2022, VANDERMEER2022113887}. The study \cite{On_the_pareto_front} uniquely explores the concept of Pareto fronts within the PINN framework, and discusses trade-offs between competing objectives. Meanwhile, other gradient-based methods have concentrated on adaptive weighting schemes, dynamically adjusting weights across loss terms to improve training stability and convergence \cite{bishof_2021, LIU2021112, McClenny_2023, XIANG202211, basir2023adaptiveaugmentedlagrangianmethod, SON2023126424, ANAGNOSTOPOULOS2024116805, Maddu_2022, VANDERMEER2022113887}. Evolutionary algorithms offer a gradient-free alternative applicable to a broader range of deep learning tasks, including PINNs \cite{lu_nsga-pinn_2023, 4492360, Wang2011, Coello2007}. However, due to their computational expense, these methods are typically limited to smaller networks and are often employed in tasks such as neural architecture search \cite{10004638}.
}

In contrast to multiobjective deep learning, people have also started using deep learning as an alternative method to solve {\color{\revisioncolor}multiobjective optimization problems (MOPs)} \cite{Li_2021, navon2021learning}.
\subsection{Organization}
We begin in section \ref{sec:MOO} with the fundamentals of MOO, giving a brief overview on various methods and focusing on two methods in more detail, namely WS and MGDA. In section \ref{sec:MOO&Dl}, we delve into the basics of DL and explore the rationale behind employing MOO for DL tasks. Using PINNs as a case study, we frame the optimization of data and physics loss as an MOO problem. Section \ref{sec:results} is dedicated to the key pitfalls we identified in applying MOO to DL. We showcase the different pitfalls arising in the training of PINNs with MOO in varying experimental setups on three example problems of increasing difficulty: the logistic equation, the heat equation, and the Rayleigh-Benard convection. The key features of the Pareto fronts obtained are discussed in detail. Apart from highlighting common pitfalls in MOO for DL, this section also brings to light additional challenges we encountered during our research. \textcolor{\revisioncolor}{In particular, we address the following challenges in detail.\\}
\begin{itemize}
    \item \textcolor{\revisioncolor}{\textbf{Identifying the Pareto front}: We provide a quick guide to identify the right regions of the solution space as the Pareto front. Moreover, the existence of different types of Pareto fronts and the challenges associated with identifying them are also highlighted.}
    \item \textcolor{\revisioncolor}{\textbf{Check if the objectives are conflicting}: We demonstrate that MOO is only beneficial if the objectives at hand are conflicting in nature. We also critique previous studies where MOO was employed even though the objectives at hand were not conflicting intrinsically.}
    \item \textcolor{\revisioncolor}{\textbf{Considering the right scaling}: To know the true curvature of a Pareto front, it is important to visualize a Pareto front on a linearly scaled plot. A Pareto front which is convex on a linear scale might appear as non-convex on a non-linear scale which can lead to wrong conclusions.}
    \item \textcolor{\revisioncolor}{\textbf{Know your optimization method}: The choice of the optimization method is crucial in order to obtain the correct Pareto front efficiently. For example, while the WS method is easy to implement, it is not capable of locating the non-convex parts of the Pareto front. Furthermore, some methods are known to scale poorly with higher dimensions. An example for this are evolutionary algorithms like NSGA-II.}
    \item \textcolor{\revisioncolor}{\textbf{Convergence}: In order to obtain the true Pareto front, we should make sure that the optimization method that we use converges. We should also make sure that our model does not underfit or overfit.}
\end{itemize}

Finally, we conclude our study by summarizing our findings in section \ref{sec:conclusion}.

%% file: chapters_post_feedback/theory/MOO.tex
In this section, we give a brief introduction to MOO. In-depth discussions can be found in, e.g., \cite{Miettinen1998, Ehrgott2005}. 
\subsection{Foundations of multiobjective optimization}
\label{subsec:multiobjective_optimiaztion}
The goal of MOO is to minimize multiple objective functions. Opposing classical literature, we use the notation $\Loss_i$ instead of $f_i$ to denote objective functions in this paper, since we are interested in the minimization of losses in the context of machine learning. The decision variables which are the trainable weights of a deep neural network are denoted by $\boldsymbol{\theta}$ since the variable $\mathbf{x}$ is reserved for input data. The general MOO problem reads
\begin{align}
    \min_{\boldsymbol{\theta} \in \R^n}\left[ \begin{array}{c}
        \Loss_1(\boldsymbol{\theta})  \\
        \vdots \\
        \Loss_m(\boldsymbol{\theta})
    \end{array}\right],
 \label{eq:MOP}
\tag{MOP}
\end{align}
and formalizes the problem of jointly minimizing the objective functions $\Loss_i$ by choosing an appropriate decision vector $\boldsymbol{\theta} \in \R^n$. Using the vector-valued function $\mathbfcal{L}: \R^n \to \R^m, \boldsymbol{\theta} \mapsto \left( \Loss_1(\boldsymbol{\theta}),\dots, \Loss_m(\boldsymbol{\theta})\right)^{\top}$ problem \eqref{eq:MOP} can also be written as
\begin{align*}
    \min_{\boldsymbol{\theta} \in \R^n} \mathbfcal{L}(\boldsymbol{\theta}).
\end{align*}
The space of inputs $\R^n$ is called the \emph{decision space} and the space $\R^m$ is called the \emph{image space}. The vectors $\boldsymbol{\theta} \in \R^n$ are \emph{decision vectors}. We call the vectors $\mathbfcal{L}(\boldsymbol{\theta}) \in \R^m$ the objective values and the set $\mathbfcal{L}(\R^n) \subseteq \R^m$ the attainable set. In many settings, it is desirable to constrain the decision space to a subset $\boldsymbol{\Theta} \subseteq \R^n$. However, we only consider the unconstrained case here, as this is most common in deep learning.\\ \\
Up to this point, it is not clear what is meant by solving \eqref{eq:MOP}. In general, there is no decision vector $\boldsymbol{\theta}^*$ that minimizes all objectives at the same time. Since there is no (natural) total order on the image space $\R^m$ we need a suitable definition of optimality for \eqref{eq:MOP}.

\input{chapters/theory/TikzFigures/DominatingPoints}

\begin{mydef}(Pareto optimality and Pareto set)
    \begin{enumerate}
        \item[i)] A decision vector $\boldsymbol{\theta}^* \in \R^n$ is (globally) \emph{Pareto optimal} if there does not exist another decision vector $\boldsymbol{\theta} \in \R^n$ such that $\Loss_i(\boldsymbol{\theta}) \le \Loss_i(\boldsymbol{\theta}^*)$ for all $i = 1,\dots,m$ and $\Loss_j(\boldsymbol{\theta}) < \Loss_j(\boldsymbol{\theta}^*)$ for some index $j \in \{1,\dots,m\}$.
        \item[ii)] The set $\mathcal{P} \coloneqq \left\lbrace \boldsymbol{\theta} \in \R^n \, : \, \boldsymbol{\theta} \text{ is Pareto optimal } \right\rbrace \subseteq \R^n$ is called the \emph{Pareto set} and the set $\mathcal{F} \coloneqq \mathbfcal{L}(\mathcal{P}) \subseteq \R^m$ is called the \emph{Pareto front}.
        \item[iii)] A decision vector $\boldsymbol{\theta}^* \in \R^n$ is \emph{locally Pareto optimal} if it satisfies i) on a neighborhood of $\boldsymbol{\theta}^*$.
    \end{enumerate}
    \label{def:Pareto_optimal}
\end{mydef}

\begin{mydef}(Weak Pareto optimality and weak Pareto set)
\begin{enumerate}
        \item[i)] A decision vector $\boldsymbol{\theta}^* \in \R^n$ is (globally) \emph{weakly Pareto optimal} if there does not exist another decision vector $\boldsymbol{\theta} \in \R^n$ such that $\Loss_i(\boldsymbol{\theta}) < \Loss_i(\boldsymbol{\theta}^*)$ for all $i = 1,\dots,m$.
        \item[ii)] The set $\mathcal{P}_w \coloneqq \left\lbrace \boldsymbol{\theta} \in \R^n \, : \, \boldsymbol{\theta}\text{ is weakly Pareto optimal } \right\rbrace \subseteq \R^n$ is called the \emph{weak Pareto set} and the set $\mathcal{F}_w \coloneqq \mathbfcal{L}(\mathcal{P}_w) \subseteq \R^m$ is called the \emph{weak Pareto front}.
        \item[iii)] A decision vector $\boldsymbol{\theta}^* \in \R^n$ is \emph{locally weakly Pareto optimal} if it satisfies i) on a neighborhood of $\boldsymbol{\theta}^*$.
    \end{enumerate}
    \label{def:weak_Pareto_optimal}
\end{mydef}

Figure \ref{fig:dominating_paretofront} provides a visualization of the definition of Pareto optimal points. The Figure shows a schematic plot of the objective function values for an MOP with two objectives. The gray area is the attainable set. The green boundary on the lower left is the Pareto front. The elements $\boldsymbol{\theta}_2$ and $\boldsymbol{\theta}_3$ are Pareto optimal since there are no other objective function values to the lower left of $\mathbfcal{L}(\boldsymbol{\theta}_2)$ and $\mathbfcal{L}(\boldsymbol{\theta}_3)$ in the sense of Definition \ref{def:Pareto_optimal}. On the other hand, $\boldsymbol{\theta}_1$ is not Pareto optimal since $\mathbfcal{L}(\boldsymbol{\theta}_2) < \mathbfcal{L}(\boldsymbol{\theta}_1)$ which is clarified by the cone starting in $\mathbfcal{L}(\boldsymbol{\theta}_1)$. In this setting, we say that decision vector $\boldsymbol{\theta}_2$ \emph{dominates} $\boldsymbol{\theta}_1$. We use the same terminology for the objective vectors, i.e., $\mathbfcal{L}(\boldsymbol{\theta}_2)$ dominates $\mathbfcal{L}(\boldsymbol{\theta}_1)$.

For practical applications, it is impossible to verify whether a given point satisfies the conditions in Definitions \ref{def:Pareto_optimal} and \ref{def:weak_Pareto_optimal} directly since one would have to compare infinitely many points. If the objective functions $\Loss_i$ are continuously differentiable we can use the following necessary condition for optimality which generalizes the \emph{Karush-Kuhn-Tucker conditions} (KKT) from scalar optimization (i.e., $\nabla \mathcal{L}(\boldsymbol{\theta}^*) = 0$) to MOO.

\begin{mydef}(Pareto critical points)
\begin{enumerate}
    \item[i)] A decision vector $\boldsymbol{\theta}^* \in \R^n$ is called Pareto critical if there exists a non-negative vector $\alpha \in \R_{\ge 0}^m$ with $\sum_{i=1}^m \alpha_i= 1$ satisfying
    \begin{align*}
        \sum_{i=1}^m \alpha_i \nabla \Loss_i(\boldsymbol{\theta}^*) = 0.
    \end{align*}
    \item[ii)] The set $\mathcal{P}_c \coloneqq \left\lbrace \boldsymbol{\theta} \in \R^n \, : \, \boldsymbol{\theta} \text{ is Pareto critical } \right\rbrace \subseteq \R^n$ is called the Pareto critical set.
\end{enumerate}
\label{def:pareto_critical}
\end{mydef}

From a numerical point of view, it is not straightforward to work with Definitions \ref{def:Pareto_optimal} and \ref{def:weak_Pareto_optimal}. In this setting, the definition of Pareto critical points is advantageous as it is easier to define algorithms which compute critical points. Computing only critical points is justified by the following observation. Any (weakly) Pareto optimal solution to \eqref{eq:MOP} is Pareto critical, i.e., Definition \ref{def:pareto_critical} formulates a necessary condition for Pareto optimality, as do the standard KKT conditions for $m=1$.

For a general problem \eqref{eq:MOP} the Pareto set $\mathcal{P}$, the weak Pareto set $\mathcal{P}_w$ and the Pareto critical set $\mathcal{P}_c$ consist of infinitely many points. Computing and representing these sets numerically is not possible in a straightforward manner. One could argue that from a practical point of view, it is not necessary to compute the entire Pareto set $\mathcal{P}$, as an accurate finite approximation of $\mathcal{P}$ and $\mathcal{F}$ is sufficient.

\subsection{Weighted sum method}
\label{subsec:weighted_sum}
\input{chapters/theory/TikzFigures/WeightedSumParetoFront}
Throughout this paper, we mainly focus on the WS method. This is a special instance of so-called \emph{scalarization methods}, which follow the idea of turning \eqref{eq:MOP} into a family of SOO problems. An optimal solution to the SOO problem should give a Pareto optimal solution to \eqref{eq:MOP}. Define the positive unit simplex $\Delta^m \coloneqq \left \lbrace \alpha \in \R_{\ge 0}^m \, : \, \sum_{i=1}^m \alpha_i = 1 \right \rbrace$. Given a weighting vector $\alpha \in \Delta^m$, we define the problem
\begin{align}
    \min_{\boldsymbol{\theta} \in \R^n} \sum_{i=1}^m \alpha_i \Loss_i(\boldsymbol{\theta}).
\label{eq:MOP_weighted_sum}
\tag{WS}
\end{align}

The approach of transferring the problem \eqref{eq:MOP} into the scalar problem \eqref{eq:MOP_weighted_sum} is popular. An optimal solution to \eqref{eq:MOP_weighted_sum} will give a Pareto critical solution to \eqref{eq:MOP}, since any critical point of \eqref{eq:MOP_weighted_sum} will be Pareto critical. In fact an even stronger result holds. It can be shown that any (locally) optimal solution $\boldsymbol{\theta}^*$ to \eqref{eq:MOP_weighted_sum} is a (locally) weakly Pareto optimal solution for \eqref{eq:MOP} \cite{Eichfelder2008}.

There is an intuitive relation between \eqref{eq:MOP} and its scalarization \eqref{eq:MOP_weighted_sum} which is presented in Figure \ref{fig:weighted_sum_3_par_front}. In Figures \ref{fig:(a)} - \ref{fig:(c)} we visualize the vector $\alpha \in \Delta^m$ in image space $\R^m$ and show the contour of the attainable set $\mathbfcal{L}(\R^n)$ in red and black. Solving \eqref{eq:MOP_weighted_sum} is equivalent to pushing the hyperplane orthogonal to $\alpha$ as far as possible in the direction of $-\alpha$. In the well-behaved convex setting in Figure \ref{fig:(a)} we end up in a point where the hyperplane is tangential to the set $\mathbfcal{L}(\R^n)$. In the non-convex setting in Figure \ref{fig:(b)} problems with this approach arise, as the $\alpha$ does no longer relate to a unique point on the front. As a consequence, we cannot find the non-convex part of the front, which is marked in red. Figure \ref{fig:(c)} describes the worst case, where only the two black points can be attained using the WS method.

When we use the WS method to approximate the Pareto set, we first have to choose a fixed number of weights $\left\lbrace \alpha^1, \dots, \alpha^N \right\rbrace \subset \Delta^m$ and then solve the problem \eqref{eq:MOP_weighted_sum} for all $\alpha \in \left\lbrace \alpha^1, \dots, \alpha^N \right\rbrace$ to compute optimal solutions $\boldsymbol{\theta}^j \in \argmin_{\boldsymbol{\theta} \in \R^n} \sum_{i=1}^m \alpha^j_i \Loss_i(\boldsymbol{\theta})$ for $j = 1,\dots, N$. Since any optimal solution to \eqref{eq:MOP_weighted_sum} yields a weakly Pareto optimal solution to \eqref{eq:MOP}, we know that $\tilde{\mathcal{P}} \coloneqq \left\lbrace \boldsymbol{\theta}^1, \dots, \boldsymbol{\theta}^N\right\rbrace \subset \mathcal{P}_w$. Therefore, $\tilde{\mathcal{P}}$ can be seen as an approximation of the weak Pareto set $\mathcal{P}_w$. A good approximation consists of decision vectors $\boldsymbol{\theta}^j$ which are either evenly distributed in the decision space or in the image space. This is a challenge since for a general \eqref{eq:MOP} it is not clear how to choose the weights $\alpha^j$ \textit{a priori} to find evenly distributed solutions $\tilde{\mathcal{P}}$. Since we have to solve \eqref{eq:MOP_weighted_sum} for every $\alpha^j$, poorly chosen weights result in computational overhead. Additionally, one has to keep in mind how the WS method scales with the number of objective functions. The simplex $\Delta^m$ has $m-1$ dimensions. Therefore, the size of a uniform discretization of the space of weights $\Delta^m$ scales exponentially with the number of objective functions in \eqref{eq:MOP}. 

\subsection{Multiobjective gradient descent algorithm}
\label{subsec:MGDA}
The \emph{multiobjective gradient descent algorithm} (MGDA) introduced in \cite{Mukai1980, Fliege2000} is an iterative method that computes from an initial point $\boldsymbol{\theta}_0$ a sequence
\begin{align}
\label{eq:general_iterative_method}
    \boldsymbol{\theta}_{k+1} = \boldsymbol{\theta}_k + \eta_k v_k, \,\text{ for }\, k \ge 0,
\end{align}
given step size $\eta_k > 0$ and directions $v_k$. The step size and direction should be chosen in a way such that \eqref{eq:general_iterative_method} yields a descent for all objective functions, i.e., $\Loss_i(\boldsymbol{\theta}_{k+1}) < \Loss_i(\boldsymbol{\theta}_k)$ for all $k \ge 0$ and all $i = 1,\dots,m$. In SOO, we can use the negative gradient as a step direction and compute a suitable step size by different strategies (e.g., constant step size, backtracking, diminishing step size). In MOO this is not possible since we have $m$ gradients $\nabla \Loss_i(\boldsymbol{\theta}_k)$. Fortunately, there is a way to compute a common descent direction from these gradients using a subroutine. Defining
\begin{align}
\label{eq:mo_steepest_descent_direction}
    v_k \coloneqq \argmin_{v \in \R^n} \max_{i=1,\dots,m} \langle \nabla \Loss_i(\boldsymbol{\theta}_k), v \rangle + \frac{1}{2}\lVert v \rVert^2,
\end{align}
there exists a step size $\eta_k > 0$ such that $\Loss_i(\boldsymbol{\theta}_{k+1}) < \Loss_i(\boldsymbol{\theta}_k)$ for all $i = 1,\dots,m$, given the vector $\boldsymbol{\theta}_k$ is not Pareto critical. It can be shown that the multiobjective steepest descent update \eqref{eq:general_iterative_method} with step direction \eqref{eq:mo_steepest_descent_direction} and a suitable step size strategy converges to Pareto critical points of \eqref{eq:MOP}. Note that problem \eqref{eq:mo_steepest_descent_direction} is computationally not too demanding since it can be reduced to a quadratic optimization problem that is $m$ dimensional (i.e., the number of objectives).\\
In general, we do not apply plain gradient descent in deep learning, but more sophisticated methods using momentum, weight decay, and stochastic gradient information, most notably the ADAM Algorithm \cite{kingma2017adam}. Similar adaptions as in the single objective setting are also possible in MOO. See, e.g., \cite{Liu2021} for stochastic MGDA and \cite{Sonntag2024, Sonntag2023} for methods with momentum.

\subsection{Brief overview of alternative methods}
\label{subsec:other_methods}
Despite the above-mentioned shortcomings, this paper focuses on the WS approach. However, there is a vast body of literature on alternative methods to find solutions to \eqref{eq:MOP} that all have their individual strengths and weaknesses.

In terms of scalarization, there are multiple methods trying to overcome the limitation of WS for non-convex problems. Examples are the $\varepsilon$-contraint method \cite{Haimes1971, Chankong1983}, the elastic constraint methods \cite{Ehrgott2002}, Benson's method \cite{Benson1978} and the reference point method \cite{Wierzbicki1980, Wierzbicki1980b}. There are further scalarization methods like Chebyshev scalarization \cite{Bowman1976, hotegni2024multiobjective}, Pascoletti-Serafini scalarization \cite{Pascoletti1984}, the normal boundary intersection method \cite{Das1998} and hybrid methods \cite{Miettinen1998, Wendell1977,Corley1980}.

Besides scalarization, there are multiple classes of algorithms that aim at directly solving \eqref{eq:MOP}. A particularly popular class is evolutionary algorithms. However, they tend to scale poorly for very large parameter dimensions such that they are of little use in the context of deep learning. One of the most popular algorithms of this class is NSGA-II \cite{Deb2002}.

In addition to the gradient methods mentioned in subsection \ref{subsec:MGDA}, there exist other first- and second-order methods. The resulting algorithms converge to single points on the Pareto set, and they are usually highly efficient. Other first-order methods are conjugate gradient methods \cite{Perez2018}, conditional gradient methods \cite{Assunccao2023} or trust-region methods \cite{Qu2013,Berkemeier2021, Berkemeier2022}. Typical higher-order methods are extensions of the Newton method to MOO, see, e.g.,  \cite{Fliege2009, Fukuda2014}. While these methods do only compute a single element of the Pareto set it is possible to adapt them in order to compute multiple solutions \cite{Bieker_Katharina_Peitz, amakor2024multiobjective}. \textcolor{\revisioncolor}{Additionally, navigation-based \cite{hartikainen2019nonconvex} methods allow decision makers to interactively explore the Pareto front, providing flexibility in adjusting preferences to understand the trade-offs between objectives in real-time.}

%% file: chapters/theory/TikzFigures/DominatingPoints.tex
\begin{figure}
\centering
\begin{tikzpicture} [line width=1, scale=1 ]
        \draw[black, thick, -stealth] (-.6  ,0.3) -- (4 ,0.3);
        \draw[black, thick, -stealth] (0-.3,-.3 + 0.3) -- (0-.3,4 + 0.3);

        \node at (3.66, -.33 + .3) {$\Loss_1$};
        \node at (-.33 - .3, 3.66 + .3) {$\Loss_2$};
        
\draw [fill=gray!40] plot [smooth cycle, tension = .5] coordinates {
(1.5, 4) (0.79, 3.71) (.6, 3) (0.79, 2.29) (1.5, 2.1) (1.9, 1.9) (2.1, 1.5) (2.29 + .1, 0.79 + .3) (3, .6 + .3) (3.71, 0.79 + .3) (4, 1.5) (4,3) (3.71, 3.71) (3, 4)
} node at (3.2,3.3)[ scale=1] {$\mathbfcal{L}(\R^n)$};

\begin{scope} 
    \clip (.79, 3.71) circle[radius=10mm];
    \clip (0-.3, 0) rectangle (4.3,4);
    \fill[nearly transparent, black] (0-.3,0) rectangle (.79, 3.71); 
\end{scope}

\begin{scope}
 \clip (-.3,0.3) rectangle (3,3);
\draw [color = black!30!green, line width=0.4mm] plot [smooth cycle, tension = .5] coordinates {
(1.5, 4) (0.79, 3.71) (.6, 3) (0.79, 2.29) (1.5, 2.1) (1.9, 1.9) (2.1, 1.5) (2.29 + .1, 0.79 + .3) (3, .6 + .3) (3.71, 0.79 + .3) (4, 1.5) (4,3) (3.71, 3.71) (3, 4)
};
\end{scope}
\fill[black!30!green] (.6, 3) circle (.08);
\fill[black!30!green] (3, .9) circle (.08);

\begin{scope} 
    \clip (1.9, 1.9) circle[radius=10mm];
    \draw (0,0) rectangle (1.9, 1.9) [color = white!0, fill = black!30!green!20];
    \draw [black!30!green, line width=0.4mm, densely dotted] (1.9, 1.9) -- (1.9, .3);
    \draw [black!30!green, line width=0.4mm, densely dotted] (1.9, 1.9) -- (0, 1.9);
\end{scope}

\fill[black!30!green] (1.9, 1.9) circle (.08);
\draw node at (2.3, 2.15)[scale=1, color = black!30!green] {$\mathbfcal{L}(\theta_3)$};

\draw [black, line width=0.4mm, densely dotted] (.79, 3.71) -- (.79, 2.71);
\draw [black, line width=0.4mm, densely dotted] (.79, 3.71) -- (-0.21, 3.71);

\fill[black] (.79, 3.71) circle (.08);
\draw node at (.79-.32, 3.71+.3)[scale=1, color=black] {$\mathbfcal{L}(\theta_1)$};
\draw node at (.15, 2.75)[scale=1, color = black!30!green] {$\mathbfcal{L}(\theta_2)$};

\end{tikzpicture}
\caption{Attainable set, Pareto front and different objective values for an abstract example \eqref{eq:MOP}.}
        \label{fig:dominating_paretofront}
\end{figure}
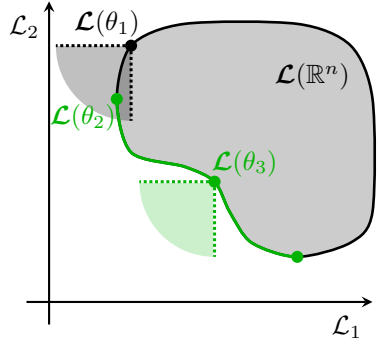

%% file: chapters/theory/TikzFigures/WeightedSumParetoFront.tex
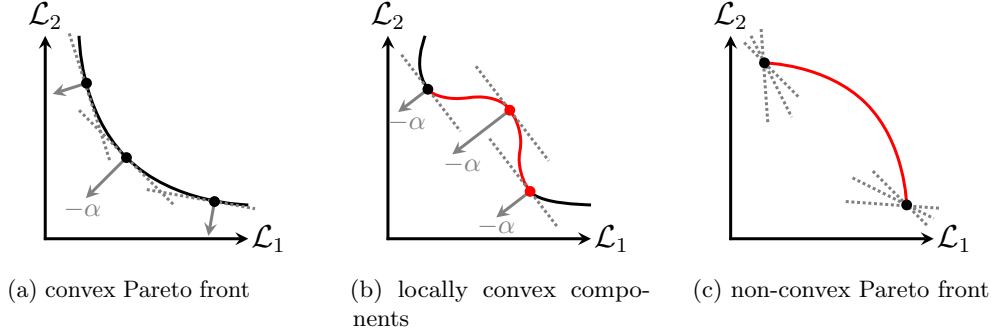
\begin{figure}
     \begin{subfigure}[b]{0.3\textwidth}
         \centering
         \begin{tikzpicture}[scale=.9]
        
            \draw[black, line width=0.4mm, -stealth] (-.02,0) -- (3,0);
            \draw[black, line width=0.4mm, -stealth] (0,-.02) -- (0,3);
    
            \node at (3.3, 0) {\large{$\Loss_1$}};
            \node at (0, 3.3) {\large{$\Loss_2$}};
    
            \draw [black, line width=0.4mm] plot [smooth, tension=1] coordinates { (.5,3) (1.2,1.2) (3,.5)};
    
                \draw [gray, line width=0.4mm, densely dotted] (0.605 - 0.25,2.25 - 0.25*-3.14) -- (0.605 + 0.35, 2.25 + 0.35*-3.14);
                \draw [gray, line width=0.4mm, -stealth] (0.61, 2.3) -- (0.61 - .5, 2.3 -.5/3.14);
                \fill (0.61,2.3) circle (0.08);

                \draw [gray, line width=0.4mm, densely dotted] (2.5 - 1, 0.55 - 1*-0.177) -- (2.5 + 0.6, 0.55 + 0.6*-0.177);
                \draw [gray, line width=0.4mm, -stealth] (2.5, 0.55) -- (2.5 - .09, 0.55 -.09/0.177);
                \fill (2.5,0.55) circle (0.08);

                \draw [gray, line width=0.4mm, densely dotted] (1.2 - 0.7, 1.2 - 0.7*-1) -- (1.2 + 0.7, 1.2 + 0.7*-1);
                \draw [gray, line width=0.4mm, -stealth] (1.2, 1.2) -- (1.2 - .6, 1.2 -.6);
                \fill (1.2,1.2) circle (0.08);
                \node at (.55,.44)  {{\color{gray}$-\alpha$}};
            
        \end{tikzpicture}
         \caption{convex Pareto front {\color{white} filler filler}}
         \label{fig:(a)}
    \end{subfigure}
    \hfill
    \begin{subfigure}[b]{0.3\textwidth}
        \centering
        \begin{tikzpicture}[scale=.9]
        
            \draw[black, line width=0.4mm, -stealth] (-.02,0) -- (3,0);
            \draw[black, line width=0.4mm, -stealth] (0,-.02) -- (0,3);
    
            \node at (3.3, 0) {\large{$\Loss_1$}};
            \node at (0, 3.3) {\large{$\Loss_2$}};

            
            \draw [black, line width=0.4mm] plot [smooth, tension=1] coordinates {(.55, 3) 
                        (.47, 2.55)
                        (.6, 2.2)}; 

            \draw [red, line width=0.4mm] plot [smooth, tension=1] coordinates {(.6, 2.2) 
                         (.9, 2.08)
                         (1.5, 2.08)
                        (1.8, 1.9)};

            \draw [red, line width=0.4mm] plot [smooth, tension=1] coordinates {(1.8, 1.9) 
                        (1.95, 1.55)
                        (1.95, 1.05)
                        (2.1, .7)};

            \draw [black, line width=0.4mm] plot [smooth, tension=1] coordinates {(2.1, .7)
                        (2.4, .56)
                        (3, .5)};   

            \draw [gray, line width=0.4mm, densely dotted] (0.6 - 0.4, 2.2 - 0.4*-1.3) -- (.6 + 0.47, 2.2 + 0.47*-1.3);
            \draw [gray, line width=0.4mm, densely dotted] (1.8 - 0.5, 1.9 - 0.5*-1.3) -- (1.8 + 0.575, 1.9 + 0.575*-1.3);
            \draw [gray, line width=0.4mm, densely dotted] (2.1 - 0.6, .7 - 0.6*-1.3) -- (2.1 + 0.4, .7 + 0.4*-1.3);

            \draw [gray, line width=0.4mm, -stealth] (.59, 2.21) -- (.59-.44, 2.21-.44/1.3);
            \draw [gray, line width=0.4mm, -stealth] (1.8, 1.9) -- (1.8-.85, 1.9-.85/1.3);
            \draw [gray, line width=0.4mm, -stealth] (2.1, .7) -- (2.1-.5, .7-.5/1.3);

            \fill (.59, 2.21) circle (.08);
            \fill[red] (1.8, 1.9) circle (.08);
            \fill[red] (2.1, .7) circle (.08);

                \node at (.28, 1.72)  {{\color{gray}$-\alpha$}};
                \node at (1.1, 1.1)  {{\color{gray}$-\alpha$}};
                \node at (1.6, .18)  {{\color{gray}$-\alpha$}};   
            
        \end{tikzpicture}
         \caption{locally convex components}
         \label{fig:(b)}
     \end{subfigure}
    \hfill
     \begin{subfigure}[b]{0.3\textwidth}
         \centering
         \begin{tikzpicture}[scale=.9]
        
            \draw[black, line width=0.4mm, -stealth] (-.02,0) -- (3,0);
            \draw[black, line width=0.4mm, -stealth] (0,-.02) -- (0,3);
    
            \node at (3.3, 0) {\large{$\Loss_1$}};
            \node at (0, 3.3) {\large{$\Loss_2$}};
    
            \draw [red, line width=0.4mm] plot [smooth, tension=1] coordinates { (.5, 2.6) (2, 2) (2.6, .5)};

                \draw [gray, line width=0.4mm, densely dotted] (.5 - .3, 2.6 + .3) -- (.5 + .5, 2.6 - .5);
                \draw [gray, line width=0.4mm, densely dotted] (.5 - .2, 2.6 + .4) -- (.5 + .4, 2.6 -.8);
                \draw [gray, line width=0.4mm, densely dotted] (.5 - 0.05, 2.6 + .5) -- (.5 + 0.05, 2.6 - .9);
                \fill (.5, 2.6) circle (.08);

                \draw [gray, line width=0.4mm, densely dotted] (2.6 - .5, .5 + .5) -- (2.6 + .3, .5 - .3);
                \draw [gray, line width=0.4mm, densely dotted] (2.6 -.8, .5 + .4) -- (2.6 +.4, .5 -.2);
                \draw [gray, line width=0.4mm, densely dotted] (2.6 -.9, .5 +0.05) -- (2.6 +.5 , .5 -0.05);
                \fill (2.6, .5) circle (.08);
                
        \end{tikzpicture}
         \caption{non-convex Pareto front {\color{white} filler}}
         \label{fig:(c)}
     \end{subfigure}
        \caption{Schematic Pareto fronts for varying degrees of convexity. The red parts indicate the sections of the Pareto front that can not be reached using the WS method (compare \cite[Figure 1]{Bieker_Katharina_Peitz}).}
        \label{fig:weighted_sum_3_par_front}
\end{figure}

%% file: chapters_post_feedback/theory/MOOFDL.tex
At the heart of DL \cite{Goodfellow-et-al-2016, BishopB24} lies a neuron, a computational unit inspired by the human brain. A network of interconnected neurons in a layer-like fashion is called an artificial neural network (ANN). A fundamental property of ANNs is their capacity to approximate any continuous function \cite{universal_csanad_csaji_approximation_2001} to arbitrary precision given a sufficiently large network architecture. A neuron processes a $d$-dimensional vector \textbf{z} as input and produces a scalar $a_i(\mathbf{z})$. The operation $a_i(\mathbf{z})$ consists of a linear transformation followed by a nonlinear one known as the activation function, 
\input{chapters/theory/TikzFigures/DNN}
\begin{equation}
    a_i(\mathbf{z}) = \sigma_i \big( \mathbf{w}^{(i)}(\mathbf{z})  + b^{(i)}\big),
\end{equation}
where $\mathbf{w}^{(i)} = \big( w_1^{(i)}, w_2^{(i)}, \cdots w_d^{(i)} \big)$ and $b^{(i)}$ correspond to the weights and the bias of the $i_{th}$ neuron, respectively. Together, we denote the parameters for the $i_{th}$ neuron as $\theta^{(i)} = \big( b^{(i)}, \mathbf{w}^{(i)} \big)$. Deep Neural Networks (DNNs) are constructed by stacking multiple layers of neurons together, which are typically divided into three types: the input layer, the hidden layers, and the output layer as shown in Figure \ref{fig:DNN}. Indexing all neurons from all layers with $i$ we denote the model parameters as $\bm{\theta}$, consisting of the concatenation of all $\theta^{(i)}$. 

Given a dataset $D = \{(\mathbf{z}_1, u_1), (\mathbf{z}_2, u_2), \cdots , (\mathbf{z}_\mathrm{N}, u_\mathrm{N})\}$ where $\mathbf{z}_i \in \R^d$ are input vectors and $u_i \in \R $ are the corresponding labels, the goal in supervised learning is to estimate a predictor $\hat{u}(\cdot, \bm{\theta^*}): \R^d \longrightarrow \R$ with learned parameters $\bm{\theta^*}$ such that $\hat{u}(\mathbf{z}_i, \bm{\theta^*})\approx u_i$. The essence of \textit{learning} entails finding $\bm{\theta^*}$ to match the input-output behavior on the training set $D$ as accurately as possible while ensuring that the learned pattern generalizes to unseen samples. The learning process in DL occurs iteratively, updating the parameters $\boldsymbol{\theta}$ in each step to minimize the discrepancy between the predicted outputs and the actual labels. Mathematically, this translates to minimizing a loss function $\Loss(\hat{u}(\mathbf{z}_i, \boldsymbol{\theta}), u_i)$ which encodes the difference between the model prediction $\hat{u}(\mathbf{z}_i, \bm{\theta})$ and the known label $u_i$. A very common objective function to minimize is the \textit{Mean Squared Error}.    
\begin{equation}
\text{MSE} = \frac{1}{N} \sum_{i=1}^{N} (\hat{u}_i - u_i)^2,
\label{eqn:mse}
\end{equation}
where $\hat{u}_i = \hat{u}(\mathbf{z}_i, \bm{\theta})$. The MSE in equation \eqref{eqn:mse} can be minimized using gradient-based methods, e.g., \textit{simple gradient descent} \cite{nonlinearProgramming}, \textit{stochastic gradient descent} \cite{nonlinearProgramming}, etc., where the \textit{backpropagation algorithm} \cite{backpropagation} is used to compute gradients. As we are interested in optimizing multiple objectives using various MOO methods, an interesting case of conflicting objectives arises in the optimization of PINNs, particularly when dealing with noisy data.

\subsection{Physics Informed Neural Networks}
Recently, PINNs \cite{karniadakis_physics-informed_2021, 712178, raissi2017physicsinformeddeeplearning, blechschmidt2021three, cuomo2022scientific} have caught a lot of attention in the ML community. This is because PINNs present a versatile and new approach to integrating domain knowledge and data-driven methods in solving complex scientific and engineering problems. PINNs leverage the universal approximation capabilities of ANNs \cite{universal_csanad_csaji_approximation_2001} to approximate solutions to differential equations. What sets PINNs apart from traditional neural network applications is the incorporation of physical laws and governing equations in the loss function. This not only enables the neural network to learn from available data but also ensures that its predictions respect the underlying physics of the system that generates the data. With the integration of such physics-based constraints, PINNs can be effective even in data-scarce scenarios, which makes them particularly appealing for applications where data collection is expensive or challenging. Furthermore, the integration of the underlying physical laws acts as a regularization term, which is particularly useful when using noisy data. Moreover, PINNs can also integrate fundamental physical principles, such as conservation laws \cite{JAGTAP2020113028, wu2021modified, baez2024guaranteeing, WU2022112143, LIN2022111053, liu2024discontinuity} and Hamiltonian dynamics \cite{MORADI20235152, greydanus2019hamiltonianneuralnetworks}, alongside domain-specific knowledge. PINNs have been deployed to solve different kinds of PDEs, e.g., integer-order PDEs \cite{raissi_physics-informed_2019}, fractional PDEs \cite{fPINNS}, integro-differential equations \cite{integro_diff_pinns}, parametric PDEs \cite{parametric_pdepinns} or stochastic PDEs \cite{Stochastic_pdepinns}. In the following, we give a brief introduction to the approach of PINNs. For a detailed overview, it is recommended to explore the work in \cite{karniadakis_physics-informed_2021}. In general, a PDE can be represented as: 
\begin{equation}
\begin{split}
    \mathcal{N}[u(\mathbf{x}, t)] + u_t(\mathbf{x}, t) &= 0, \hspace{1.2cm}\mathbf{x} \in \Omega, t \in [0, T], \\
    u(\mathbf{x}, t_0) &= g_0(\mathbf{x}),  \hspace{.6cm} \mathbf{x} \in \Omega, \\
    u(\mathbf{x}, t) &= g_{\partial \Omega}(\mathbf{x}, t), \hspace{.1cm} \mathbf{x} \in \partial\Omega,t \in [0, T],
\end{split}
\label{eqn:generalDifferentialEquation}
\end{equation}
 where $\mathcal{N}[\cdot]$ is a nonlinear differential operator and $\Omega \subset \mathbb{R}^d$ is the spatial domain with boundary $\partial \Omega$. The functions $g_0(\mathbf{x})$ and $g_{\partial \Omega}(\mathbf{x})$ describe the initial condition (IC) and the boundary condition (BC), respectively. 

The solution to a PDE is a function that depends on space or, in the time-dependent case, on both space and time. Thus, it is a natural idea to use a DNN to represent the PDE solution via a feed-forward network. The network takes as input $\mathbf{z} = (\mathbf{x}, t)$, where $\mathbf{x} = (x_1, x_2, \dots, x_d)$ and $t$ are the spatial and temporal coordinates, respectively, and produces the solution $\hat{u}(\mathbf{x}, t; \boldsymbol{\theta})$. Assuming such a solution exists, one can define a residual function, $r(\mathbf{x}, t; \boldsymbol{\theta})$, from equation \eqref{eqn:generalDifferentialEquation}
\begin{equation*}
    r(\mathbf{x}, t; \boldsymbol{\theta}) := \mathcal{N}[\hat{u}(\mathbf{x}, t; \boldsymbol{\theta})] + \hat{u}_t(\mathbf{x}, t; \boldsymbol{\theta}),
    \label{eqn:residue}
\end{equation*}
which is ideally equal to zero, if we have the the right set of parameters $\boldsymbol{\theta}$ to represent the solution of the respective differential equation. This is achieved by encoding $r(\mathbf{x}, t; \boldsymbol{\theta})$ as a loss function called $\Loss_\mathrm{PDE}$ and using an optimization algorithm to minimize it. The ability to calculate the partial derivatives of the neural network output $\hat{u}(\mathbf{x}, t; \boldsymbol{\theta})$ with respect to the inputs (i.e., position $\mathbf{x}$ and time $t$) through backpropagation facilitates the evaluation of the partial differential operator $\mathcal{N}$ within the neural network setting. 
Following the definition of MSE from equation \eqref{eqn:mse}, one can write:
\begin{equation}
    \Loss_\mathrm{PDE}(\mathbf{x}, t, \boldsymbol{\theta}) = \frac{1}{N_\mathrm{PDE}} \sum_{j = 1} ^{N_\mathrm{PDE}}\lvert r (x_j, t_j;\boldsymbol{\theta}) \rvert^2,
    \label{eqn:pdeLoss}
\end{equation}
where $N_\mathrm{PDE}$ is the number of data points ($x_j, t_j$) used to calculate $\Loss_\mathrm{PDE}$. The points are called \textit{collocation points} and correspond to specific locations in the domain where $\Loss_\mathrm{PDE}$ is evaluated. Hence, by choosing the collocation points strategically and minimizing $\Loss_\mathrm{PDE}$, we are enforcing the neural network to satisfy the underlying PDEs at these locations. However, so far we have only encoded the residue $r(\mathbf{x}, t; \boldsymbol{\theta})$ as a loss function. To obtain a particular solution, we also need to encode the IC and the BC from equation \eqref{eqn:generalDifferentialEquation} as individual losses to be minimized:
\begin{align*}
    \Loss_\mathrm{IC}(\mathbf{x}^\mathrm{IC}, t = t_0, \boldsymbol{\theta}) &= \frac{1}{N_\mathrm{IC}} \sum_{j = 1} ^{N_\mathrm{IC}}\lvert \hat{u}(x_j^\mathrm{IC}, t_0; \boldsymbol{\theta}) - g_0(x_j^\mathrm{IC}) \rvert^2 , \\
    \Loss_\mathrm{BC}(\mathbf{x}^\mathrm{BC}, t, \boldsymbol{\theta}) &= \frac{1}{N_\mathrm{BC}} \sum_{j = 1} ^{N_\mathrm{BC}}\lvert \hat{u}(x_j^\mathrm{BC}, t_j;\boldsymbol{\theta}) - g_{\partial \Omega}(x_j^\mathrm{BC}, t) \rvert^2 ,
\end{align*}
where $\mathbf{x}^\mathrm{IC}$ and $\mathbf{x}^\mathrm{BC} \in \partial \Omega$ are the collocation points for the IC and BC, respectively. $N_\mathrm{IC}$ and $N_\mathrm{BC}$ are the number of points sampled to calculate $\Loss_\mathrm{IC}$ and $\Loss_\mathrm{BC}$, respectively. Combining $\Loss_\mathrm{PDE}$, $\Loss_\mathrm{IC}$ and $\Loss_\mathrm{BC}$ yields the physics loss $\Loss_\mathrm{PHYSICS}$
\begin{align}
   \Loss_\mathrm{PHYSICS} = \alpha_\mathrm{PDE}\Loss_\mathrm{PDE} + \alpha_\mathrm{IC}\Loss_\mathrm{IC} + \alpha_\mathrm{BC}\Loss_\mathrm{BC} ,
   \label{eqn:physicsLoss}
\end{align}
where $\alpha_\mathrm{PDE}$, $\alpha_\mathrm{IC}$, $\alpha_\mathrm{BC} \geq 0$ are the weights given to the respective losses. 

It should be noted that so far, we have not used any labeled data in defining $\Loss_\mathrm{PDE}$ in equation \eqref{eqn:physicsLoss}. If we additionally have access to a data set of measurements (e.g., from numerical simulations or experiments), then we can define the data loss, which simply is a standard loss function as we know it from supervised learning.   
\begin{equation}
    \Loss_\mathrm{DATA}(\mathbf{x}, t, \boldsymbol{\theta}) = \frac{1}{N_\mathrm{DATA}} \sum_{j = 1}^{N_\mathrm{DATA}} \lvert \hat{u}(x_j, t_j;\boldsymbol{\theta}) - \hat{u}_j \rvert^2.
    \label{eqn:dataLoss}
\end{equation}
The total loss function $\Loss(\boldsymbol{\theta})$ is a linear combination of $\Loss_\mathrm{PDE}$ and $\Loss_\mathrm{DATA}$ as follows:
\begin{equation}
    \Loss (\boldsymbol{\theta}) = \alpha_\mathrm{DATA}\Loss_\mathrm{DATA}(\boldsymbol{\theta})+ \alpha_\mathrm{PHYSICS}\Loss_\mathrm{PHYSICS}(\boldsymbol{\theta}) ,
    \label{eqn:totalLoss}
\end{equation}
where $\alpha_\mathrm{DATA}, \alpha_\mathrm{PHYSICS} \geq 0$ are the weights given to the data loss and the physics loss respectively. 

By tuning these weights, one can control the interplay between the data and the physics loss and hence affect the trainability and prediction of PINNs. However, choosing the weights is not straightforward and most of the time, either the weights are set to one ($\alpha_\mathrm{DATA}=\alpha_\mathrm{PHYSICS}=1$) or they are selected through trial and error to achieve the desired neural network predictions. In either case, it is unclear whether a particular weight is the best choice and why a specific weight might outperform others in terms of the predictive capabilities. 
\subsection{Multiobjective optimization of PINNs}
\label{subsec:multi-ob-op-pinns}
To understand how weight choices influence the optimization landscape and independent losses, one needs to rely on the multiobjective treatment of minimizing $\Loss_\mathrm{DATA}$ and $\Loss_\mathrm{PHYSICS}$ as discussed in section \ref{sec:MOO}. Hence, the problem of minimizing both losses jointly can be reformulated as an MOP 
\begin{align*}
    \min_{\boldsymbol{\theta} \in \R^n}\left[ \begin{array}{c}
        \Loss_\mathrm{DATA}(\boldsymbol{\theta})  \\
        \Loss_\mathrm{PHYSICS}(\boldsymbol{\theta})
    \end{array}\right].
 \label{eq:MOP_PINNS}
\end{align*}
We will use the WS method (subsection \ref{subsec:weighted_sum}) and MGDA (subsection \ref{subsec:MGDA}) to solve the MOP and thus find the Pareto front. For the WS method, we denote $\alpha_\mathrm{DATA}$ by $\alpha$ and $\alpha_\mathrm{PHYSICS}$ by $1 -\alpha$ in equation \eqref{eqn:totalLoss}. Additionally, we compare our results with NSGA-II \cite{Deb2002}, an evolutionary algorithm considered a benchmark in MOO. 
\subsection{Experimental Setup}
In the following, we lay out the experimental settings to use the MOO of PINNs for three equations, the logistic equation, heat equation and the Rayleigh-Bénard convection. We thus start with an ordinary differential equation, as it allows us to visualize some key aspects, before moving on to PDEs. 
\subsubsection{Logistic Equation}
\label{sec:logisitc_equation}
 The logistic equation serves as a mathematical model for population growth over time, incorporating both the population's reproductive capacity and the growth limitations imposed by resource availability, competition, and other environmental factors. This equation finds widespread application in ecology, biology, and other disciplines to investigate population dynamics. The logistics equation is expressed as follows:
\begin{equation*}
    \frac{du(t)}{dt} = r \, u(t)\left(1 - \frac{u(t)}{K}\right),
\label{eqn:logistic}
\end{equation*}
where $r$ is the intrinsic growth rate of the population and $K$ is the carrying capacity, which denotes the maximum population size that the environment can support without exceeding its resources. For simplicity, we will choose $K = 1$, and $u(0) = \frac{1}{2}$ as the IC. For this problem, there exists an exact analytic solution
\begin{align}
    u_\mathrm{exact}(t) &= \frac{1}{1+ \exp(-rt)}.
    \label{eqn:logisticexact}
\end{align}
\subsubsection{Heat Equation}
The heat equation serves as a mathematical model for the time-dependent distribution of heat in a given region. In one dimension (e.g., a rod of length $L$, where $\Omega = (0, L)$), the heat equation is written as:
\begin{equation*}
    \frac{\partial u(x, t)}{\partial t} =  \kappa \frac{\partial^2 u(x, t)}{\partial x^2}, 
    \label{eqn: heatequation}
\end{equation*}
where $\kappa$ is a positive constant called thermal diffusivity. We choose $u(x, 0) = \sin \left( \pi \frac{x}{L} \right)$ $ \forall$ $ x \in (0, L)$ as an IC, and $u(0, t) = u(L, t) = 0$ $\forall$ $t \in [0, T]$ as Dirichlet BC, respectively. With these conditions, the heat equation is well-posed with the following analytic solution:
\begin{equation}
    u_\mathrm{exact} = \sin \left( \pi \frac{x}{L} \right) \cdot e^{\frac{-\kappa \pi^2}{^2}t}.
    \label{eqn:heatsolutionexact}
\end{equation}

\color{\revisioncolor}
\subsubsection{Rayleigh-Bénard Convection}
The Rayleigh-Bénard convection (RBC) is a system of PDEs that describes the fluid and temperature flow in a narrow horizontal layer heated from below and cooled from above.
This self-organizing nonlinear system is determined by the governing PDEs

\begin{equation}
\begin{split}
\frac{\partial \mathbf{u}}{\partial t} + \mathbf{u} \cdot \nabla \mathbf{u} &= -\nabla p + T \mathbf{\hat{x}} + \sqrt{\frac{\text{Pr}}{\text{Ra}}} \nabla^2 \mathbf{u},\\[2pt]
\nabla \cdot \mathbf{u} &= 0,\\[2pt]
\frac{\partial T}{\partial t} + \mathbf{u} \cdot \nabla T &= \frac{1}{\sqrt{\text{Pr} \, \text{Ra}}} \nabla^2 T.\\
\end{split}
\end{equation}
This system creates a non-trivial coupling between velocity $\mathbf{u} = (u, v)$ with a horizontal and vertical component, temperature $T$, and pressure $p$. Two non-dimensional parameters define the behavior of this system, the Rayleigh number $\mathrm{Ra}$ and the Prandtl number $\mathrm{Pr}$. We set $\mathrm{Ra} = 10^4$ and $\mathrm{Pr} = 0.71$.
We consider the two-dimensional spatial domain $\Omega$ of a rectangle with height $L_x=2$ and width $L_y=2\pi$ over time $\Omega_t=[0,60]$. The fluid is affected by gravity and buoyancy forces, with $\mathbf{\hat{x}}$ as a unit vector in vertical direction. Horizontally, we have periodic boundary conditions, $\mathbf{u}(x, 2 \pi, t) = \mathbf{u}(x, 0, t)$ and $T(x, 2 \pi, t) = T(x, 0, t)$. The upper and lower plates are isothermal, with temperatures $T_{\text{bottom}}=1$ and $T_{\text{top}}=0$. Therefore, we set the Dirichlet boundary conditions,
$T(-1, y, t) = 1$,
$T(1, y, t) =  0$ and
$\boldsymbol{u}(\pm 1, y, t) = (0, 0)$. The initial state of the system is a vertical temperature gradient, with small perturbations. 
This setup leads to convection, with two Bénard convection cells in future states. In our experiment, $\Loss_\mathrm{DATA}$ is constructed with the temperature data, which is simulated numerically with the spectral Galerkin method, using the Shenfun Python library \cite{mortensen_shenfun_2018} on a set of collocation points. 
\color{black}
\subsubsection{Experiments}
We use equations \eqref{eqn:dataLoss} and \eqref{eqn:physicsLoss} to calculate the data loss $\Loss_\mathrm{DATA}$ and the physics loss $\Loss_\mathrm{PHYSICS}$ for the logistic equation, the heat equation and the RBC problem, respectively. The task of minimizing both losses is formulated as an MOP. For the optimization, we employ both the WS and MGDA techniques. The implementation of MGDA is an adaptation from \cite{sener_multi-task_nodate} where the dual problem is solved instead of the primal\footnote{For a detailed discussion, please refer to \cite{Fliege2000}.} problem. We use the Adam optimizer (with $ \eta=0.003$) to train feed-forward networks with layer sizes \textcolor{\revisioncolor}{$(1,9,9,9,1)$ for the logistic equation, $(2,50,50,50,50,1)$ for the heat equation, and models with two different hidden layer sizes for RBC}. $\Loss_\mathrm{DATA}$ and $\Loss_\mathrm{PHYSICS}$ are calculated on the same grid of equally spaced collocation points: $20$ points for the logistic equation, $20\times20$ grid in the spatial-temporal domain of the heat equation, \textcolor{\revisioncolor}{and $32\times48\times300$ grid in the vertical-horizontal-temporal dimension for the RBC.}
The collocation points that fall on the boundaries are used to calculate the loss for the boundary conditions. We corrupt the values taken from the exact solution $u_i$ at the collocation points by adding Gaussian noise with a mean of $0$ and standard deviation $\sigma$ (which gets specified in the particular experiments) to simulate real (noisy) data. In the subsequent chapter, we discuss the major pitfalls occurring when using MOO for DL and demonstrate some aspects using our experiments. 

%% file: chapters/theory/TikzFigures/DNN.tex
\begin{figure}
    \centering
\begin{tikzpicture}[shorten >=1pt,auto,node distance=2cm,semithick, scale=0.75, transform shape]

  \newcommand{\inputNodes}{4}
  \newcommand{\hiddenOneNodes}{5}
  \newcommand{\hiddenTwoNodes}{6}
  \newcommand{\outputNodes}{3}

  \pgfmathsetmacro{\inputTop}{(\inputNodes - 1) / 2}
  \pgfmathsetmacro{\hiddenOneTop}{(\hiddenOneNodes - 1) / 2}
  \pgfmathsetmacro{\hiddenTwoTop}{(\hiddenTwoNodes - 1) / 2}
  \pgfmathsetmacro{\outputTop}{(\outputNodes - 1) / 2}

  \foreach \i in {0,...,\numexpr\inputNodes-1\relax}
    \node[draw,circle,fill=orange!50] (Input-\i) at (0,\inputTop-\i) {};

  \foreach \i in {0,...,\numexpr\hiddenOneNodes-1\relax}
    \node[draw,circle,fill=orange!50] (Hidden1-\i) at (3,\hiddenOneTop-\i) {};

  \foreach \i in {0,...,\numexpr\hiddenTwoNodes-1\relax}
    \node[draw,circle,fill=orange!50] (Hidden2-\i) at (6,\hiddenTwoTop-\i) {};

  \foreach \i in {0,...,\numexpr\outputNodes-1\relax}
    \node[draw,circle,fill=orange!50] (Output-\i) at (9,\outputTop-\i) {};

  \foreach \i in {0,...,\numexpr\inputNodes-1\relax}
    \foreach \j in {0,...,\numexpr\hiddenOneNodes-1\relax}
      \draw (Input-\i) -- (Hidden1-\j);

  \foreach \i in {0,...,\numexpr\hiddenOneNodes-1\relax}
    \foreach \j in {0,...,\numexpr\hiddenTwoNodes-1\relax}
      \draw (Hidden1-\i) -- (Hidden2-\j);

  \foreach \i in {0,...,\numexpr\hiddenTwoNodes-1\relax}
    \foreach \j in {0,...,\numexpr\outputNodes-1\relax}
      \draw (Hidden2-\i) -- (Output-\j);

  \foreach \i in {0,...,\numexpr\outputNodes-1\relax} {
      \draw[->] (Output-\i) -- ++(1,0);
  }

  \node[align=center, above] at (Input-0.north) {Input Layer};
  \node[align=center, above] at (Hidden1-0.north) {Hidden Layer \#1};
  \node[align=center, above] at (Hidden2-0.north) {Hidden Layer \#2};
  \node[align=center, above, yshift = 0.4cm] at (Output-0.north) {Output Layer};

\end{tikzpicture}
    \caption{A sketch of a DNN depicting data flow and network outputs from input to the output layer.}
    \label{fig:DNN}
\end{figure}
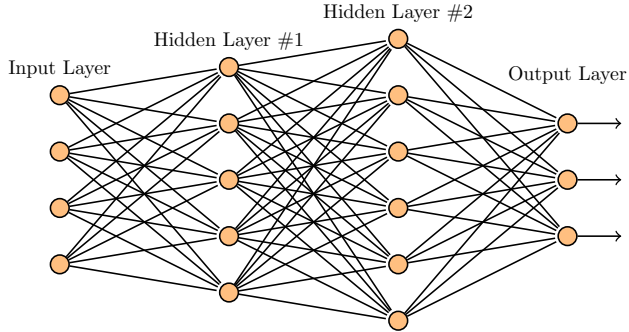

%% file: chapters_post_feedback/Results/pitfalls.tex
\textcolor{\revisioncolor}{In this chapter, we highlight challenges commonly encountered in applying MOO for DL. To illustrate these challenges, we design a series of experiments training PINNs to solve three equations of increasing complexity: the logistic equation, the heat equation, and the RBC problem. While the logistic and heat equations provide foundational insights, the RBC problem, being significantly more complex, allows us to showcase the applicability of MOO in tackling challenging real-world scenarios. The insights gained from these experiments are not restricted to PINNs but are broadly applicable to any DL problem.
}
\subsection{Identifying the Pareto front}
\label{subsec:identifyingfront}
Imagine a scenario where one uses MOO to optimize two objective functions, $\Loss_1$ and $\Loss_2$. Figure \ref{fig:types_pareto_fronts} shows four different kinds of plots to highlight the existence of different kinds of Pareto fronts. In all four figures, the horizontal and the vertical axes represent the losses $\Loss_1$ and $\Loss_2$, respectively.  The gray area is the attainable set and the green boundary is the Pareto front. \textcolor{\revisioncolor}{While in Figure \ref{fig:typesofFronts(a)}, the Pareto front exhibits a convex shape, which allows for smooth and predictable trade-offs between objectives, making it easier to identify solutions using methods like WS, Figure \ref{fig:typesofFronts(b)} displays a Pareto front with both convex and non-convex regions. The non-convex areas in Figure \ref{fig:typesofFronts(b)} are significant as they require advanced algorithms to capture solutions that might be missed by simpler methods, offering opportunities for improved trade-offs that are not apparent in purely convex scenarios. Figure \ref{fig:typesofFronts(c)} shows the existence of a discontinuous Pareto front, which represents cases where solutions cluster in isolated regions, making the optimization process more challenging and highlighting the need for methods capable of navigating abrupt shifts in trade-offs. Finally, it is also possible to have fronts that are locally Pareto optimal but are dominated globally, as described in Figure \ref{fig:typesofFronts(d)}, where the blue parts are locally Pareto optimal but get dominated by other solutions. These locally optimal regions are important as they can provide insights into region-specific trade-offs when the exploration of the global Pareto front is computationally infeasible, while the globally Pareto optimal regions, shown in green, represent the most competitive trade-offs overall.}
\input{chapters/theory/TikzFigures/TypesOfParetoFronts}
Hence, one should always keep in mind that any of the four cases could be present in one's specific application, such that it is likely possible to end up with incomplete Pareto fronts, clustered points, or only local optima. While these issues are easily visualized for low-dimensional problems, ignorance of these challenges can easily lead to poor solutions in high-dimensional problems, as in the case of DL. What adds even more to the challenge is the fact that DNN training is characterized by many local optima. In the worst case, one can even end up with a large fraction of non-optimal solutions (see, e.g., \cite{On_the_pareto_front} Figure 3). \textcolor{\revisioncolor}{Compared to MOO, there is only one optimal objective function value in SOO, so identifying trade-offs between objectives is irrelevant.}\\
\textcolor{\revisioncolor} {\paragraph{\textbf{Suggested Solution}} A straightforward fix to mitigate these challenges is to employ a \textit{non-dominance test} as a post-processing step. Therein, we eliminate all the points that are dominated by at least one other point in the identified Pareto set, meaning that it is superior in all objectives. For instance, in Figure \ref{fig:logisticandheatfronts}, the gray points represent dominated solutions that are filtered out through the non-dominance test, as they are outperformed across all objectives.} 
\subsection{Check if the objectives are conflicting}
\label{subsec:conflictingobjectives}
In section \ref{sec:MOO}, we discuss the importance of MOO. \textcolor{\revisioncolor}{However, these discussions are only meaningful if the objective functions are conflicting. We call the objective functions conflicting if they do not have a joint minimizer. If the objective functions are not conflicting, the Pareto front consists only of a single point. In this case all scalarizations in the weighted sum method have the same minimizer and there is no point in computing solutions for different weights. In this setting, the application of MOO techniques yields no additional benefit, as there are no different meaningful trade-offs to be computed. This problem can be observed in the two example problems that get introduced in the last section. For data without noise, the two objectives are essentially non-conflicting. The exact solution to the PDEs has minimal data and physics loss. If it can be expressed by the neural network with high accuracy, there is no meaningful amount of conflict in the objectives. Only when the data is corrupted by noise there is in general no function that fits the data exactly while following the physical laws.} To study this issue we have corrupted the solution data with different levels of Gaussian noise. 
\begin{figure}[htbp]
    \centering
    \begin{subfigure}{.495\textwidth}
        \includegraphics[width=\linewidth]{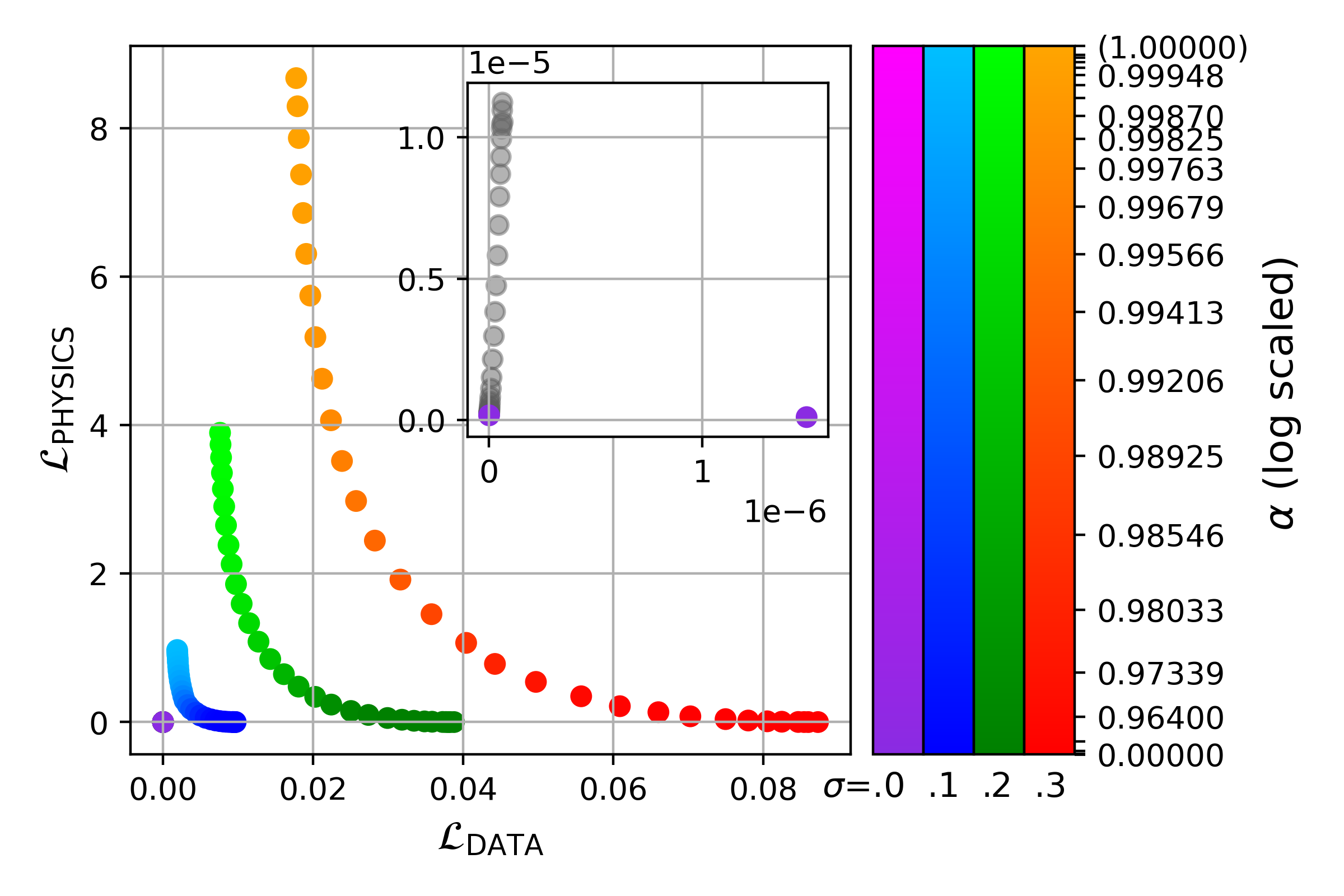}
        \caption{Logistic equation}
        \label{fig:logisticparetonoise}
    \end{subfigure}
    \hfill
    \begin{subfigure}{.495\textwidth}
        \includegraphics[width=\linewidth]{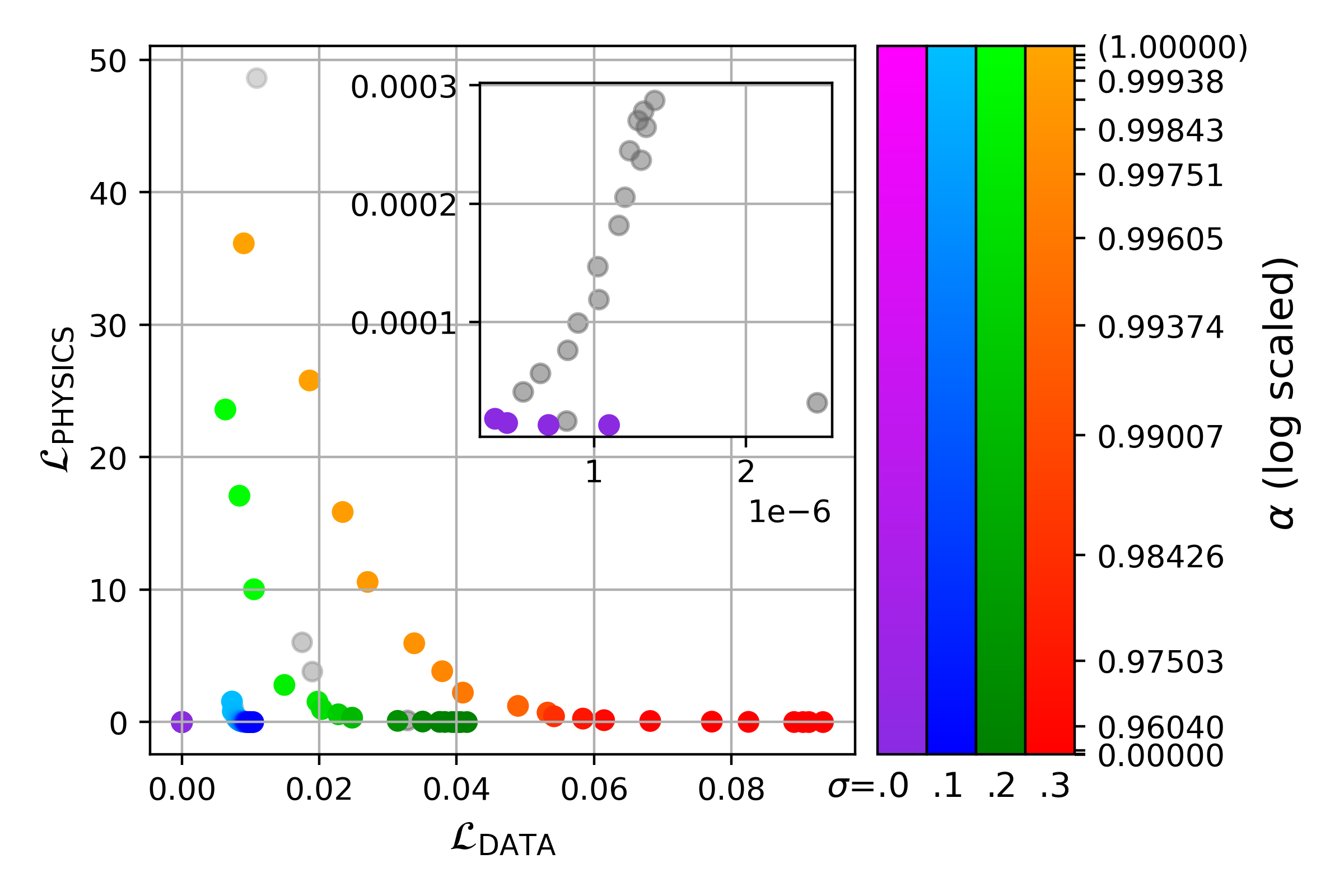}
        \caption{Heat equation}
        \label{fig:heatparetonoise}
    \end{subfigure}
    \caption{Pareto fronts for logistic and heat equation with
varying noise level $\sigma \in \{0.0, 0.1, 0.2, 0.3\}$, indicated by different colors. The total loss is $\Loss(\boldsymbol{\theta}) = \alpha \Loss_\mathrm{DATA} + (1-\alpha)\Loss_\mathrm{PHYSICS}$ with $\alpha \in [0, 1]$.}
    \label{fig:logisticandheatfronts}
\end{figure}

\begin{figure}[htbp]
    \centering
    \begin{subfigure}{.495\textwidth}
        \includegraphics[width=\linewidth]{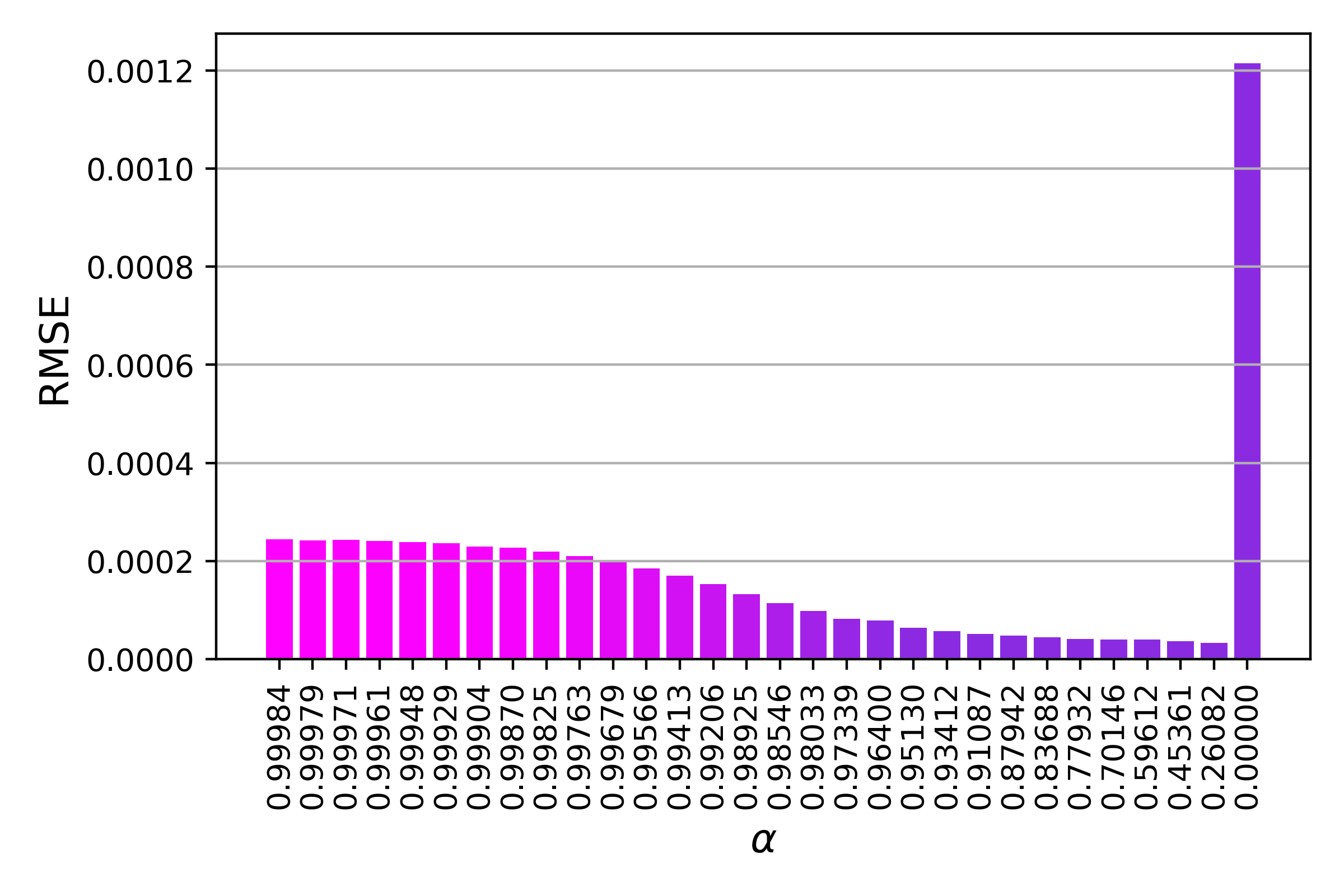}
        \caption{Without noise $(\sigma = 0.0)$}
        \label{fig:generr_logistic_sigma00}
    \end{subfigure}
    \hfill
    \begin{subfigure}{.495\textwidth}
        \includegraphics[width=\linewidth]{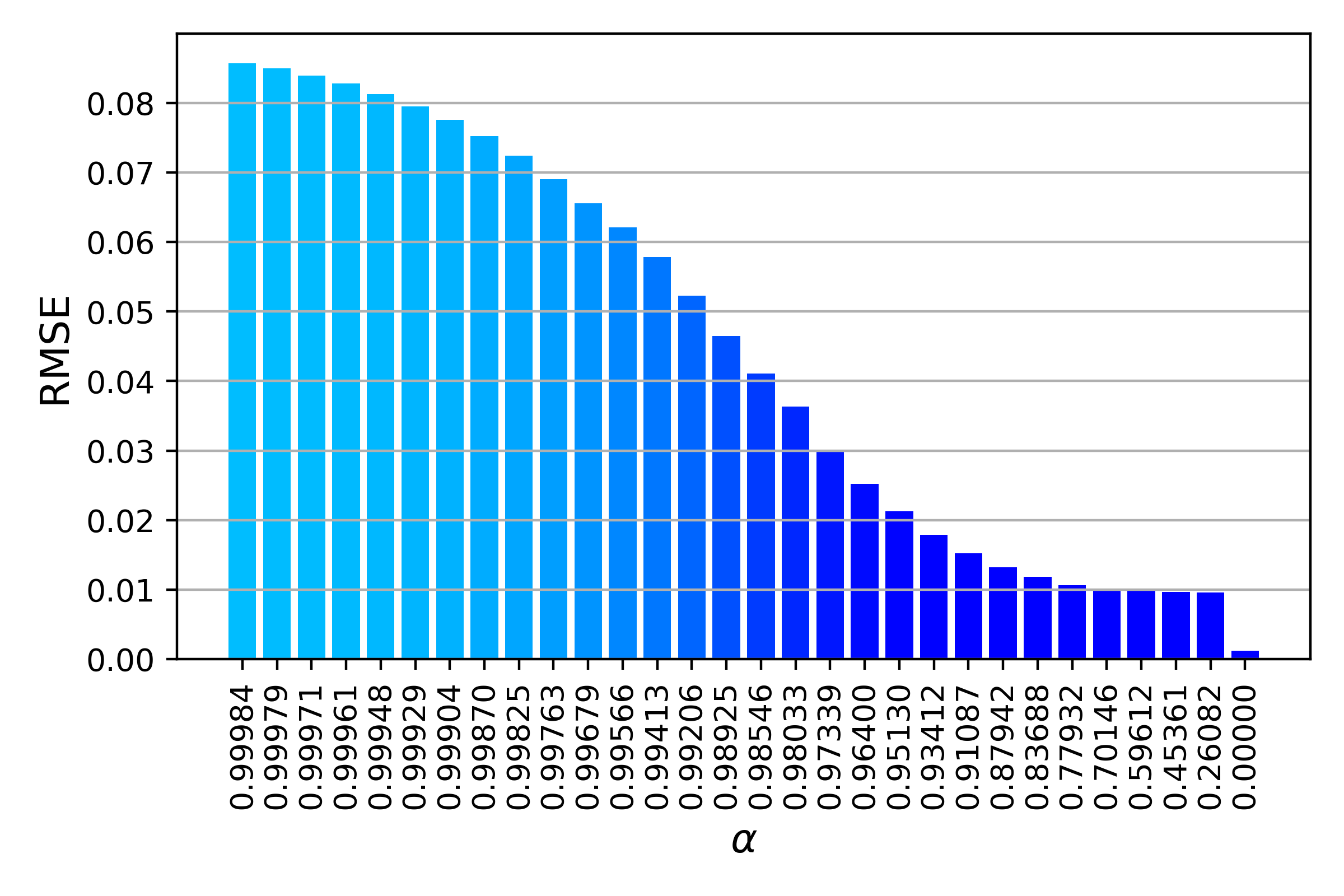}
        \caption{With noise $(\sigma = 0.1)$}
        \label{fig:generr_logistic_sigma01}
    \end{subfigure}
    \caption{Root Mean Squared Error for logistic equation models over $\alpha \in [0, 1]$. We calculate the RMSE on a densely sampled test set comparing the prediction to the analytic solution.}
    \label{fig:generalization_error}
\end{figure}

\begin{figure}[htbp]
    \centering
        \includegraphics[width=0.5\linewidth]{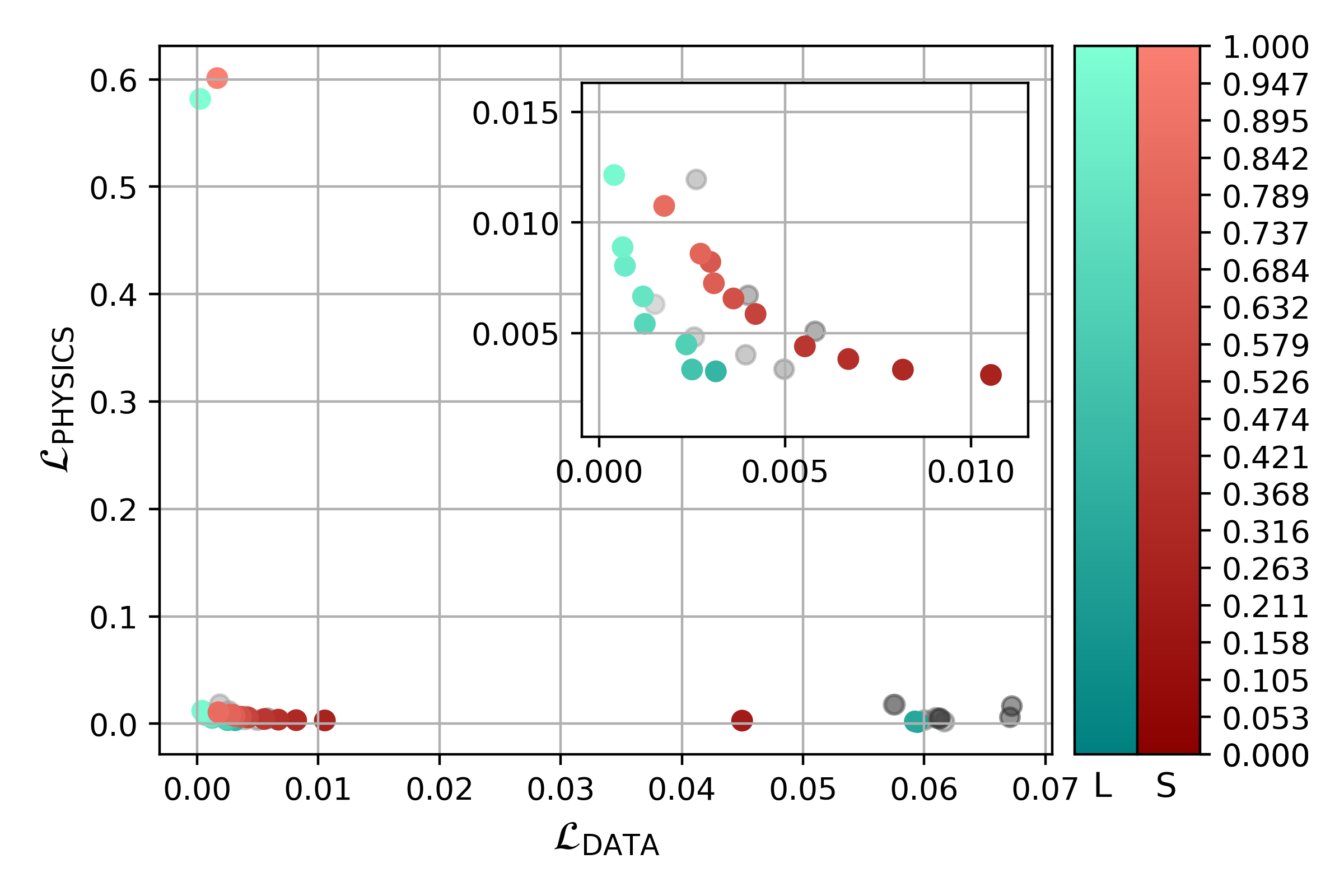}
        \caption{\textcolor{\revisioncolor}{Pareto fronts for the RBC solved with a large (L, cyan) and small (S, dark-red) DNN. The small network consists of 5 hidden layers with 50 neurons each, the large consists of 10 hidden layers with 100 neurons each.}}
        \label{fig:rbcmodels}
\end{figure}
Figures \ref{fig:logisticparetonoise} and \ref{fig:heatparetonoise} show the Pareto fronts for the logistic and the heat equation, respectively. The conflict between the objectives is signified by different Pareto fronts corresponding to increasing standard deviations of the Gaussian noise. We observe that an increased noise leads to more trade-offs between the objectives. \textcolor{\revisioncolor}{As expected, the points nearly collapse to the optimal loss value in the absence of noise (purple points in \ref{fig:logisticandheatfronts}) and no longer form a Pareto front. The small spread of purple points can be explained by numerical inaccuracies arising in the training of neural networks and the slightly different training trajectories caused by the different weighting of the two losses via $\alpha$. Figures \ref{fig:generr_logistic_sigma00} and \ref{fig:generr_logistic_sigma01} show the RMSE to the analytic solution without and with added noise, respectively. Training on noise-less data seems to be advantageous for generalization purposes. Figure~\ref{fig:generr_logistic_sigma01} shows that too much noise (here $\sigma \geq 0.1$) in the data will harm the solution with increasing weight. We expect that the optimal choice of the weight $\alpha$ with respect to generalization depends on the level of noise in the data and needs to be determined for each problem individually. For the more challenging RBC problem, we observe two distinct Pareto fronts (Figure \ref{fig:rbcmodels}), differing in the network size used for training. Unlike the logistic equation and the heat equation, no noise is added to the exact numerical solution for the RBC problem. The conflict observed in the Pareto fronts is associated with numerical noise inherent in calculating the exact numerical solution. This conflict is further magnified when using a smaller, less expressive network, which struggles to capture the underlying solution accurately. This finding highlights the critical role of network expressivity in multi-objective optimization. An inadequately expressive network can artificially induce conflicts between objectives, even when such conflicts do not inherently exist.}
\textcolor{\revisioncolor}{In contrast to these insights, previous studies, such as \cite{sener_multi-task_nodate}, employed MOO to solve multi-task problems but used highly expressive network architectures, effectively eliminating conflicts between tasks and negating the need for MOO-specific methods. Similarly, in \cite{On_the_pareto_front}, where MOO was used to calculate Pareto fronts for PINNs, no noise was introduced into the data, leading to an absence of conflict between the data and physics losses, rendering the use of MOO unnecessary. In general, to benefit from MOO, ensuring that objectives conflict is essential. This is in contrast to SOO where conflicting objectives do not exist.}\\
\textcolor{\revisioncolor} {\paragraph{\textbf{Suggested Solution}} To make sure the objectives are conflicting, one can compute the gradients of each loss with respect to the model parameters and see how these gradients align or diverge. If the gradients point in opposite directions, it suggests a conflict between the two objectives. Another solution is to run a simple MOO algorithm (e.g., NSGA-II) for a few runs to analyze if there is a trade-off in optimizing the objectives, in which case the objectives are conflicting.}
\subsection{Considering the right scaling}
\label{subsec:rightscale}
{\color{\revisioncolor} In Figure \ref{fig:types_pareto_fronts}, we observe that Pareto fronts can exhibit either a convex or non-convex shape. Convex Pareto fronts have smooth, predictable trade-offs between objectives, while non-convex Pareto fronts involve more complex trade-offs, where small changes in one objective
can lead to significant shifts in another. Importantly, the mathematical property of convexity is inherent to the objectives and does not change with the scaling of the plot. While most of the Pareto fronts in our experiments are convex, the apparent shape of the front may visually change depending on whether it is plotted on a linear or logarithmic scale. For instance, in Figure \ref{fig:heat_pareto_l5_k1_}, the Pareto front appears convex on a linear scale, but when plotted on a double logarithmic scale (Figure \ref{fig:heat_pareto_l5_k1_loglog}), it may appear non-convex due to the non-linear transformation of the axes. However, this does not affect the actual convexity of the front. 
\begin{figure}[htbp]
    \centering
    \begin{subfigure}{.495\textwidth}
        \includegraphics[width=\linewidth]{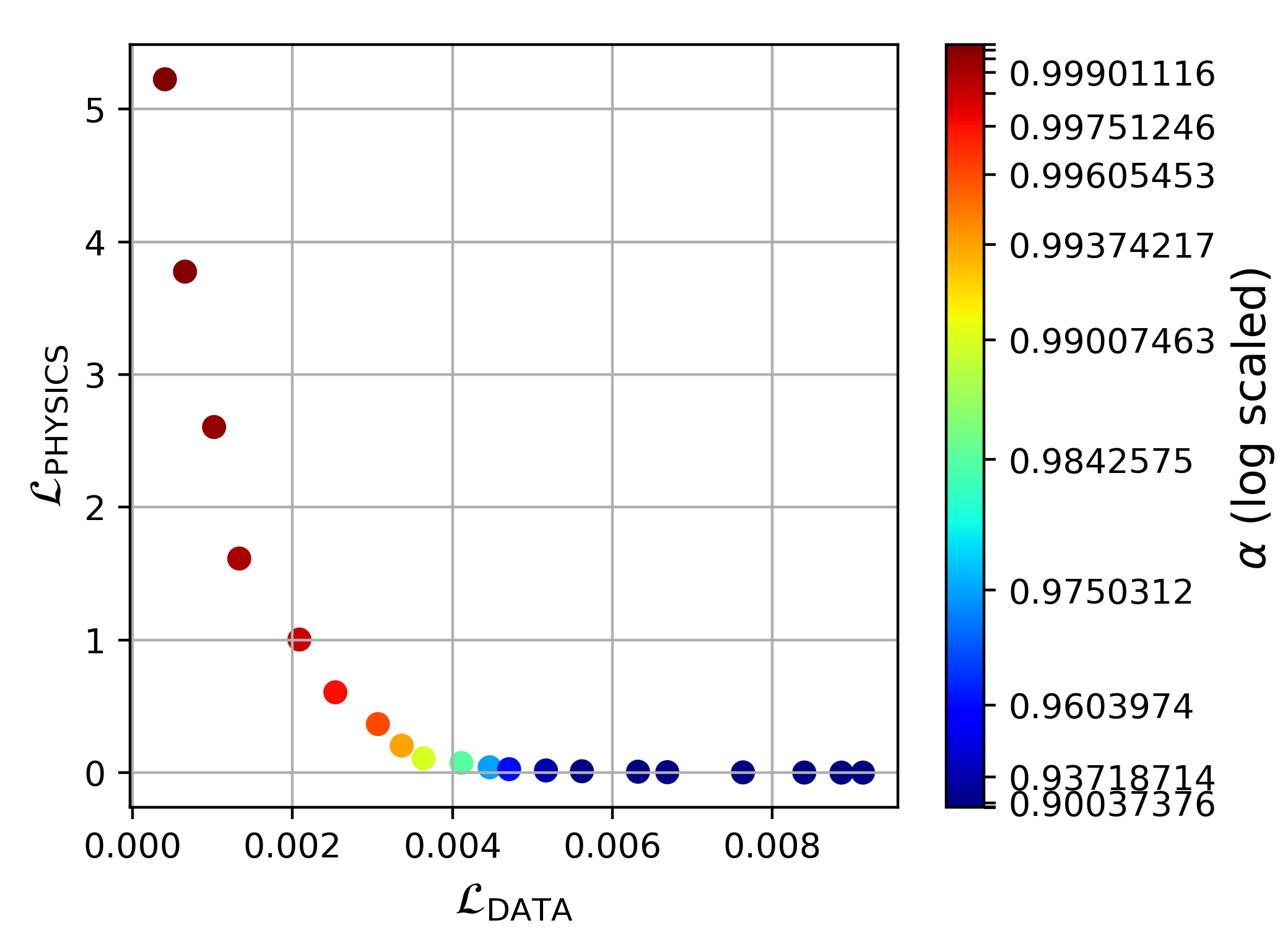}
        \caption{Pareto front on a linear scale}
        \label{fig:heat_pareto_l5_k1_}
    \end{subfigure}
    \hfill
    \begin{subfigure}{.495\textwidth}
        \includegraphics[width=\linewidth]{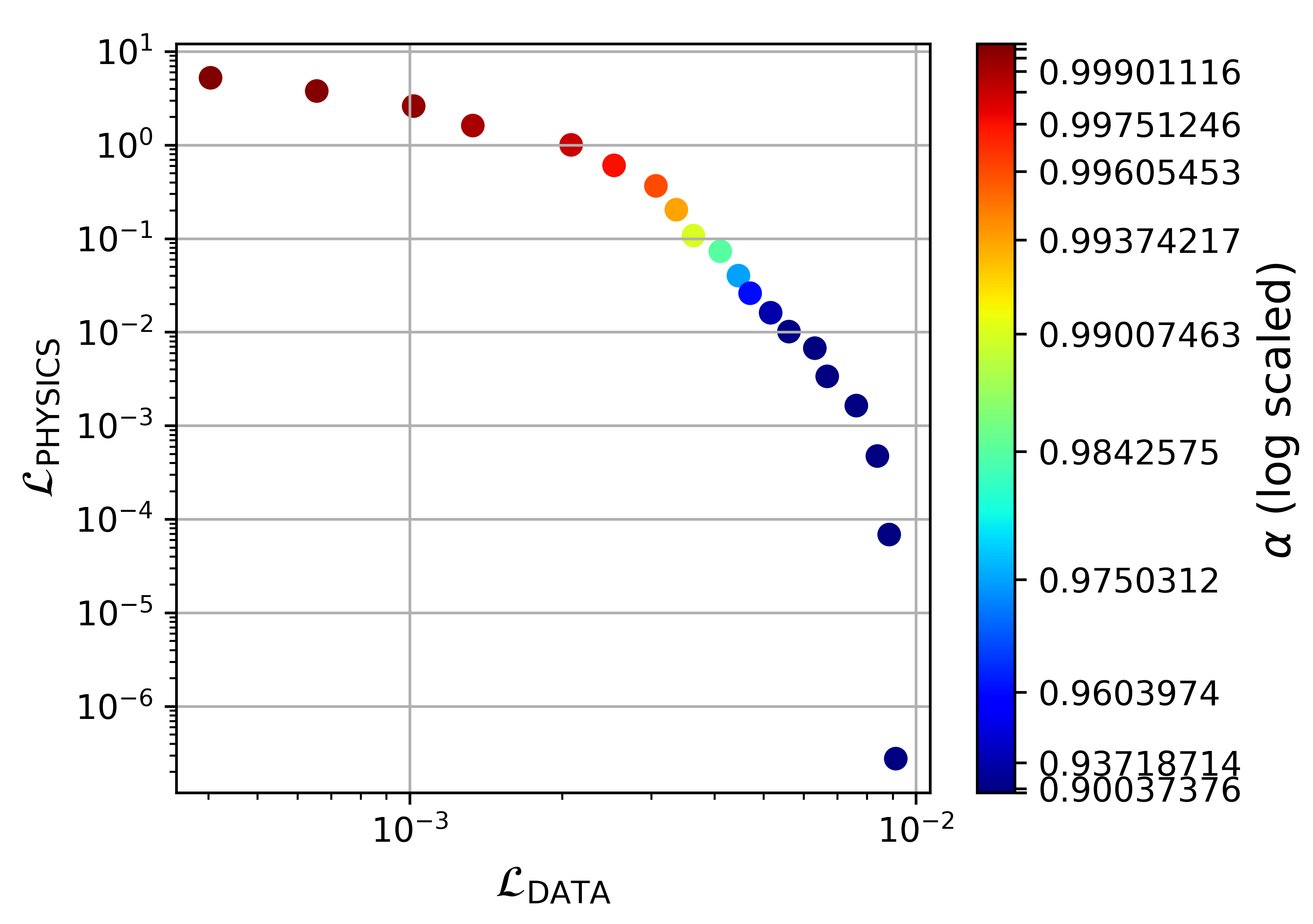}
        \caption{Pareto front on a double logarithmic scale}
        \label{fig:heat_pareto_l5_k1_loglog}
    \end{subfigure}
\caption{Pareto fronts obtained for heat equation for $L=5, k = 1$ plotted on different scales (a) normal scale, (b) double logarithmic scale}
\label{fig:loglogvsnormalheat}
\end{figure}
A notable example of the potential for deception due to scaling can be found in the study \cite{On_the_pareto_front}, where the authors used a logarithmic scale to plot the Pareto fronts. As a result, they reported non-convex regions in the front and provided different interpretations for these convex and non-convex parts. However, the non-convexity in their results was likely an artifact of the logarithmic scaling, as convexity should be evaluated on a linear-linear scale. Moreover, in this work, the authors used the weighted sum method, which in general is not capable of capturing non-convex regions of the Pareto front. This further highlights that their interpretation of non-convexity was influenced by their method and the choice of scaling, rather than the true mathematical properties of the front.
\color{\revisioncolor}\\
\paragraph{\textbf{Suggested Solution}}  To avoid misleading interpretations, we recommend testing multiple scales (e.g., linear, logarithmic) for visualization purposes but relying on the linear-linear scale when assessing convexity.}
\color{black}
\subsection{Know your optimization method}
\label{subsec:knowyourmethod}
\begin{figure}
    \centering
    \begin{subfigure}{.495\textwidth}
        \includegraphics[width =\linewidth]{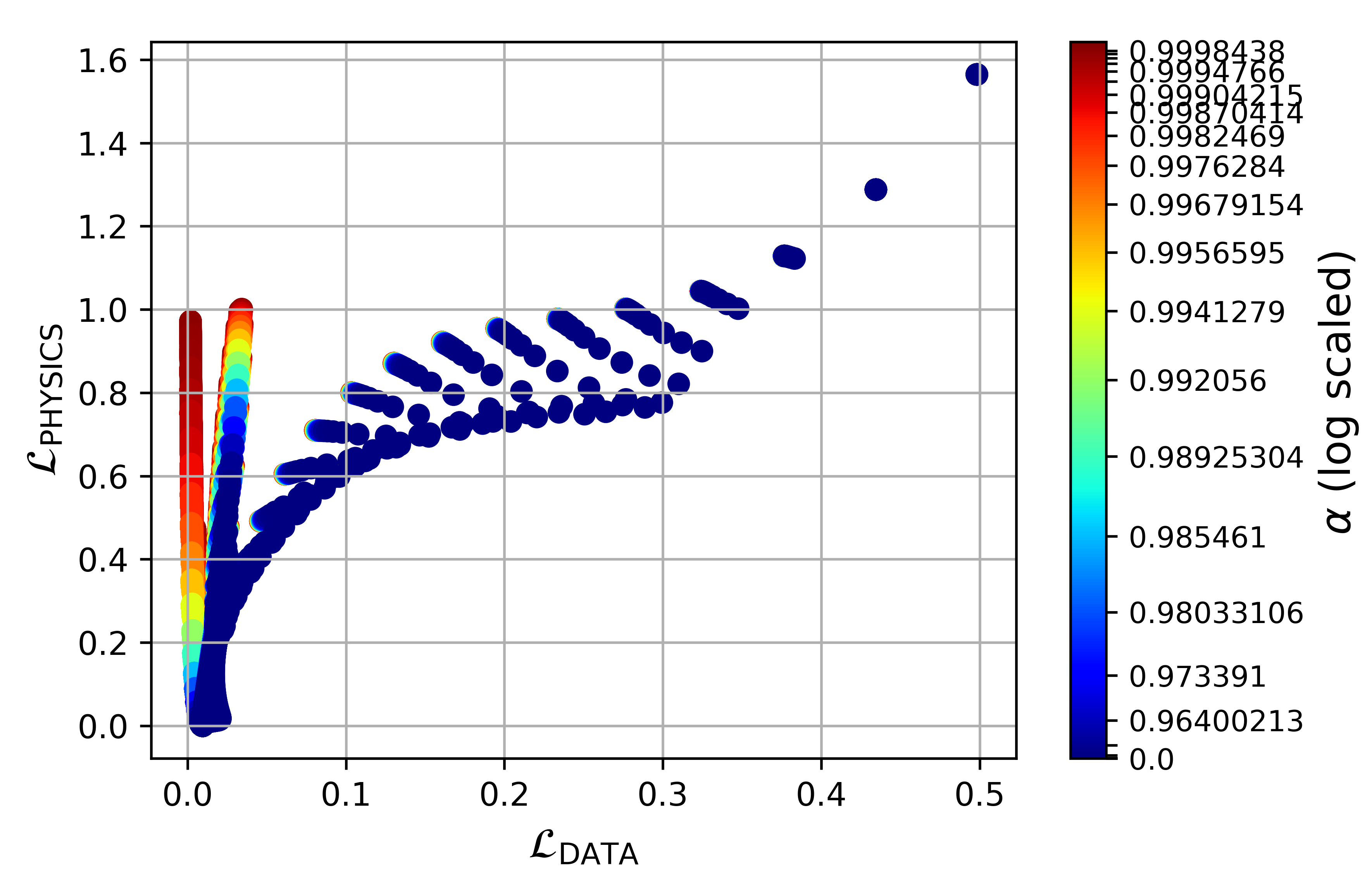}
        \caption{No epochs omitted.}
        \label{fig:log_0.1_evolutionall}
    \end{subfigure}
    \hfill
    \begin{subfigure}{.495\textwidth}
        \includegraphics[width =\linewidth]{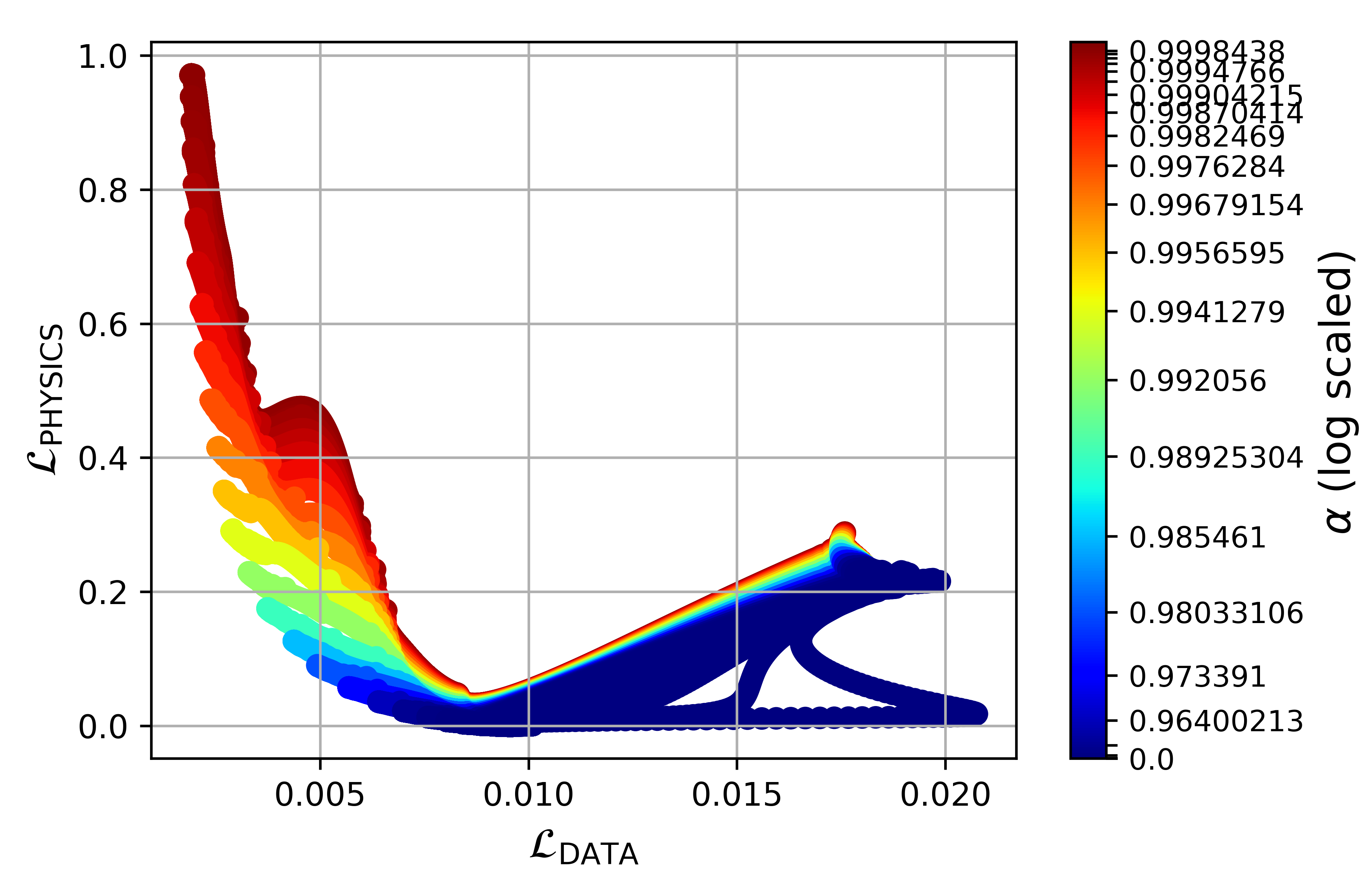}
        \caption{First 50 epochs omitted.}
        \label{fig:log_0.1_evolutionomitted}
    \end{subfigure}
\caption{Loss evolution over training epochs demonstrating the formation of the Pareto front for the logistic equation using the weighted sum method.}
\label{fig:log_0.1_evolution_plots}
\end{figure}

\begin{figure}
    \centering
    \begin{subfigure}[b]{0.495\textwidth}
        \centering
        \includegraphics[width=\linewidth]{figures/plots/heat_L5k1_pareto_normal.png}
        \caption{$L = 5, k = 1$}
        \label{fig:heat_pareto_l5_k1}
    \end{subfigure}
    \hfill
    \begin{subfigure}[b]{0.495\textwidth}
        \centering
        \includegraphics[width=\linewidth]{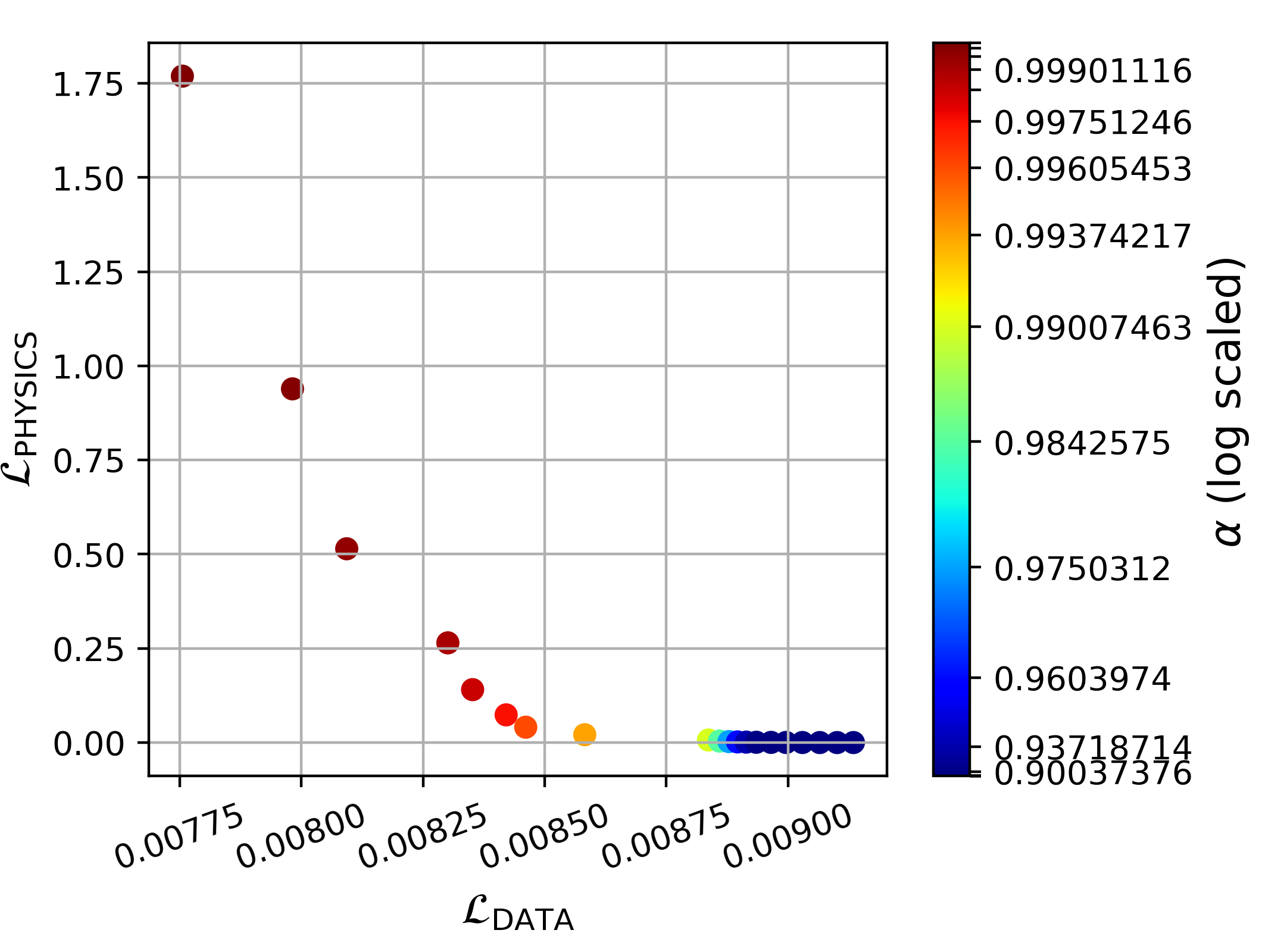}
        \caption{$L = 1, k = 1$}
        \label{fig:heat_pareto_l1_k1}
    \end{subfigure}
    \caption{Pareto fronts for heat equation with different settings using the weighted sum method.}
    \label{fig:allheatparetos}
\end{figure}

It is important to understand the optimization method one wants to employ. Especially, one should be well aware of what to expect of an optimization method. In Table \ref{tab:comparison_moo_methods}, we summarize the strengths and weaknesses of some of the most popular MOO methods. Depending on the MOP at hand, it is important to choose a suitable method. Consider the WS method, for instance, which can only locate the convex regions of a Pareto front (subsection \ref{subsec:weighted_sum}). This is consistent with the Pareto fronts that we have shown in Figure \ref{fig:logisticandheatfronts} which are all convex.
\begin{table}[h]
\centering
\caption{A summary of the strengths and weaknesses of the weighted sum (WS), Weighted Chebyshev scalarization (WCS), multiobjective gradient descent algorithm (MGDA) and evolutionary algorithms (EAs).}
\label{tab:comparison_moo_methods}
{\footnotesize 
\begin{tabular}{|p{1.3cm}|p{5.2cm}|p{5.2cm}|} 
\hline
Algorithm & Strength & Weakness \\ \hline
WS &
\vspace{-0.5em}
\begin{itemize}[leftmargin=1.0em]
\item  One can use any existing gradient descent algorithm.
\item Easy to implement.
\item One can steer the direction to specific points by parameter variation.
\end{itemize}
\vspace{-0.5em} &
\vspace{-0.5em} 
\begin{itemize}[leftmargin=1.0em]
\item Only convex regions can be explored.
\item Parametrization to obtain an equidistant distribution of points is very hard.
\item Steering towards a specific desired tradeoff is difficult.
\end{itemize}
\vspace{-0.5em} \\
\hline
WCS &
\vspace{-0.5em} 
\begin{itemize}[leftmargin=1.0em]
\item Both convex and non-convex problems can be explored.
\item One can steer the direction to specific points by parameter variation.
\end{itemize}
\vspace{-0.5em} 
&
\vspace{-0.5em} 
\begin{itemize}[leftmargin=1.0em]
\item Optimization problem with constraints $\Rightarrow$ more challenging to solve.
\item Parametrization to obtain an equidistant distribution of points is very hard.
\end{itemize}
\vspace{-0.5em} \\
\hline
MGDA &
\vspace{-0.5em} 
\begin{itemize}[leftmargin=1.0em]
\item Both convex and non-convex problems can be explored.
\item Often numerically efficient.
\item Strong theoretical foundation.
\item Parameter free.
\end{itemize}
\vspace{-0.5em} 
&
\vspace{-0.5em} 
\begin{itemize}[leftmargin=1.0em]
\item Steering the direction to specific points is not possible.
\item Multiple points on the front can be explored only via multi-start in combination with suitable initializations.
\end{itemize}
\vspace{-0.5em} \\
\hline
EAs &
\vspace{-0.5em} 
\begin{itemize}[leftmargin=1.0em]
\item Yields an approximation of the front by a population in a single run.
\item Black box treatment of objectives $\Rightarrow$ easy implementation.
\end{itemize}
\vspace{-0.5em} 
&
\vspace{-0.5em} 
\begin{itemize}[leftmargin=1.0em]
\item Computationally expensive.
\item Scale poorly w.r.t. the number of parameters.
\item No convergence guarantees for a finite number of iterations.
\end{itemize}
\vspace{-0.5em} \\
\hline
\end{tabular}
}
\end{table}
In Figure \ref{fig:log_0.1_evolutionall}, we present the evolution of the losses during training, which illustrates the formation of the Pareto front using the WS method. In this plot, for each $\alpha$, we start with the same point in the feasible space, i.e., with the same random initialization of the model, and record the losses during the training. We see that the weighted sum approach can obtain a convex Pareto front. To get a better view of the final Pareto front, we plot a zoomed-in version in Figure \ref{fig:log_0.1_evolutionomitted} by omitting the first 50 epochs. In Figures \ref{fig:heat_pareto_l5_k1} and \ref{fig:heat_pareto_l1_k1}, we show the Pareto fronts for the heat equation for different domain sizes $L$ using the WS method. As expected, irrespective of the choice of $L$, the shape of the Pareto front is convex. In contrast, the Pareto fronts shown in \cite{On_the_pareto_front} have both convex and non-convex fronts. \textcolor{\revisioncolor}{The WS method is inherently limited to convex fronts, though numerical issues can sometimes produce non-convex approximations. Consequently, the proposed relationship between Pareto front curvature and domain size is questionable. This emphasizes again the need for a careful analysis of the optimization process.}

\begin{figure}
    \centering
    \begin{subfigure}{.45\textwidth}
        \includegraphics[width =\linewidth]{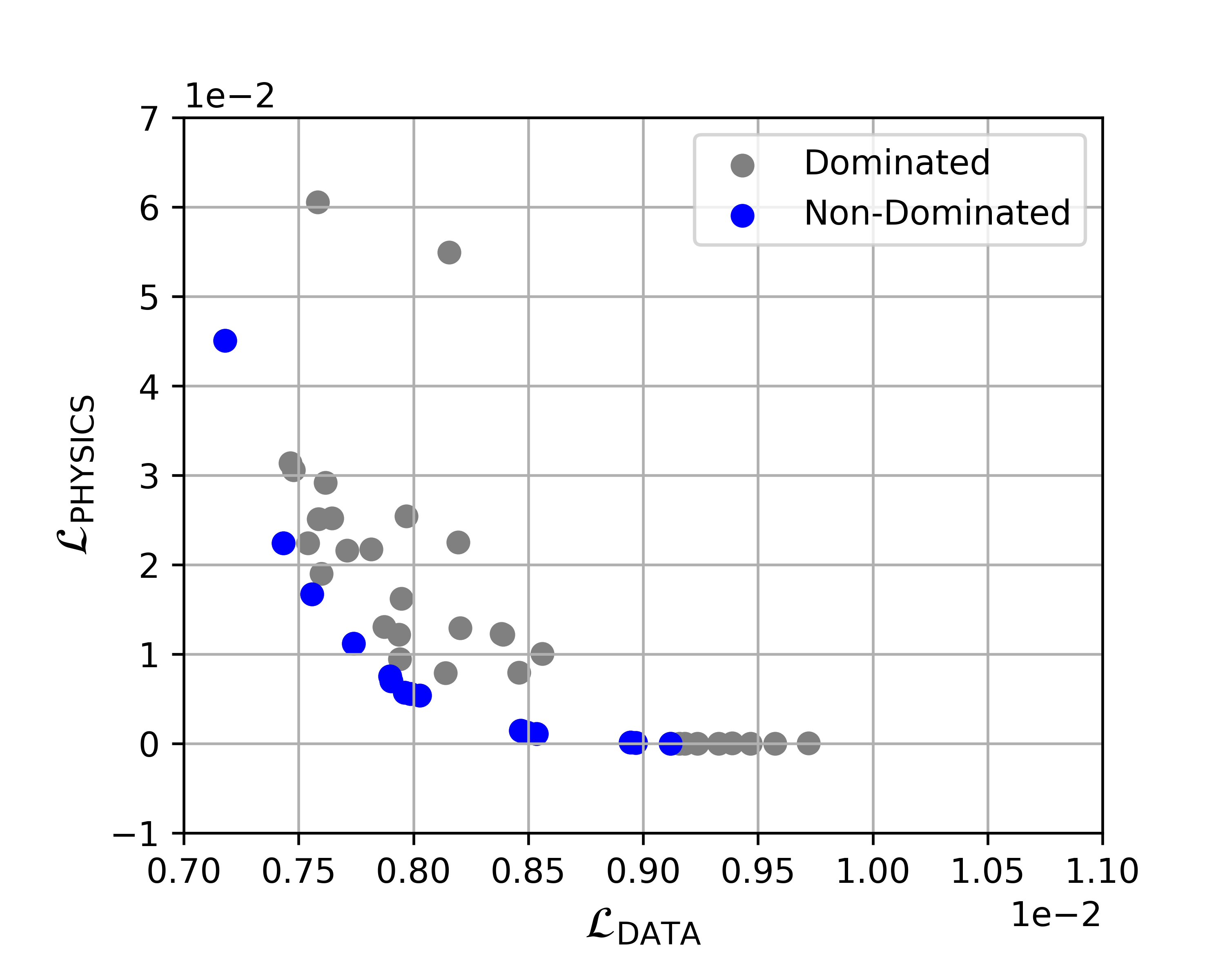}
        \caption{Gradients not normalized.}
        \label{fig:mgda_nonorm}
    \end{subfigure}
    \hfill
    \begin{subfigure}{.45\textwidth}
        \includegraphics[width =\linewidth]{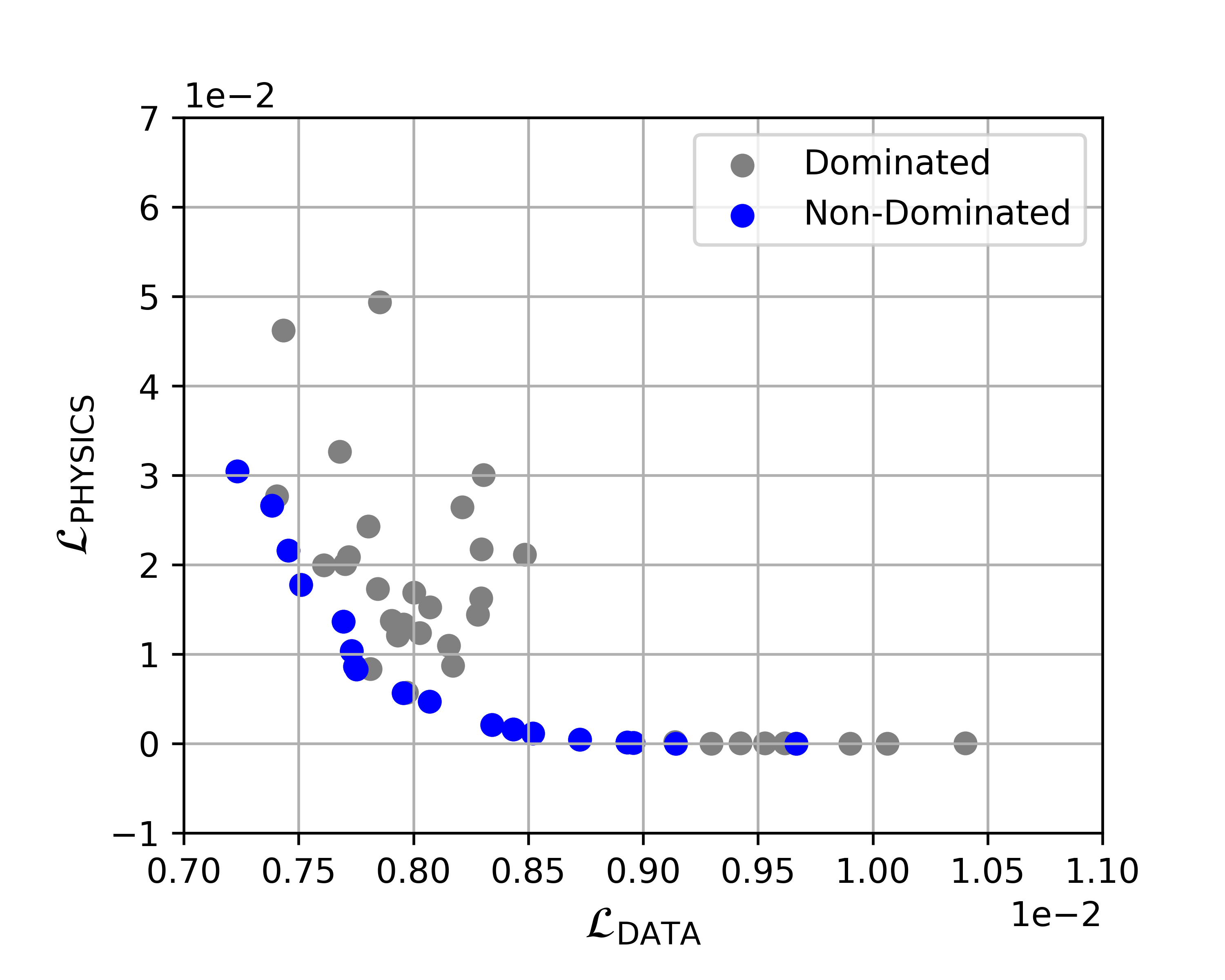}
        \caption{Gradients normalized.}
        \label{fig:mgda_norm}
    \end{subfigure}
\caption{Pareto fronts for logistic equation using MGDA.}
\label{fig:mgda_fronts}
\end{figure}

\begin{figure}
    \centering
    \begin{subfigure}[b]{0.495\textwidth}
        \centering
        \includegraphics[width=\linewidth]{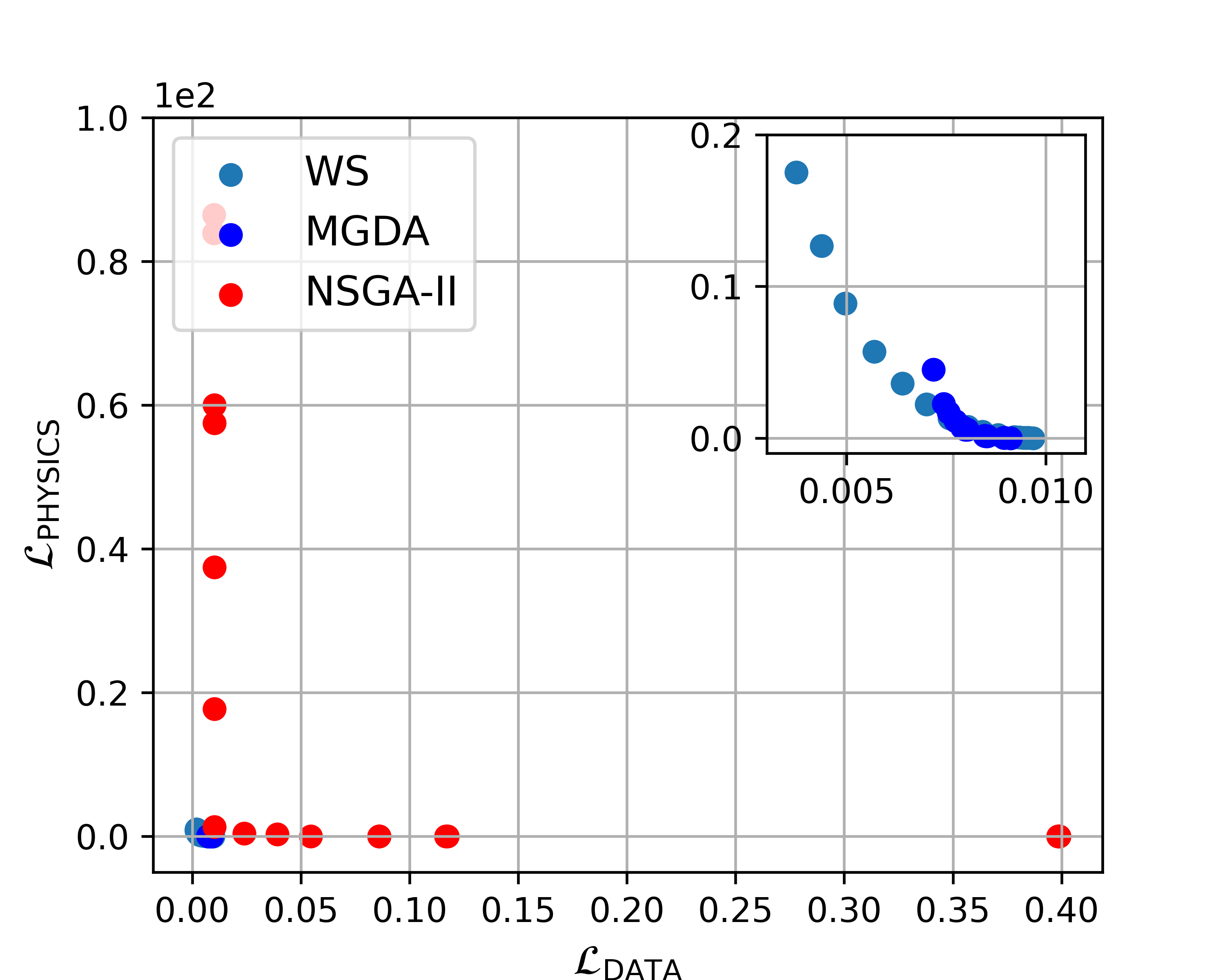}
        \caption{WS, MGDA, and NSGA-II.}
        \label{fig:ws_Vs_mgda_Vs_nsga2}
    \end{subfigure}
    \hfill
    \begin{subfigure}[b]{0.495\textwidth}
        \centering
        \includegraphics[width=\linewidth]{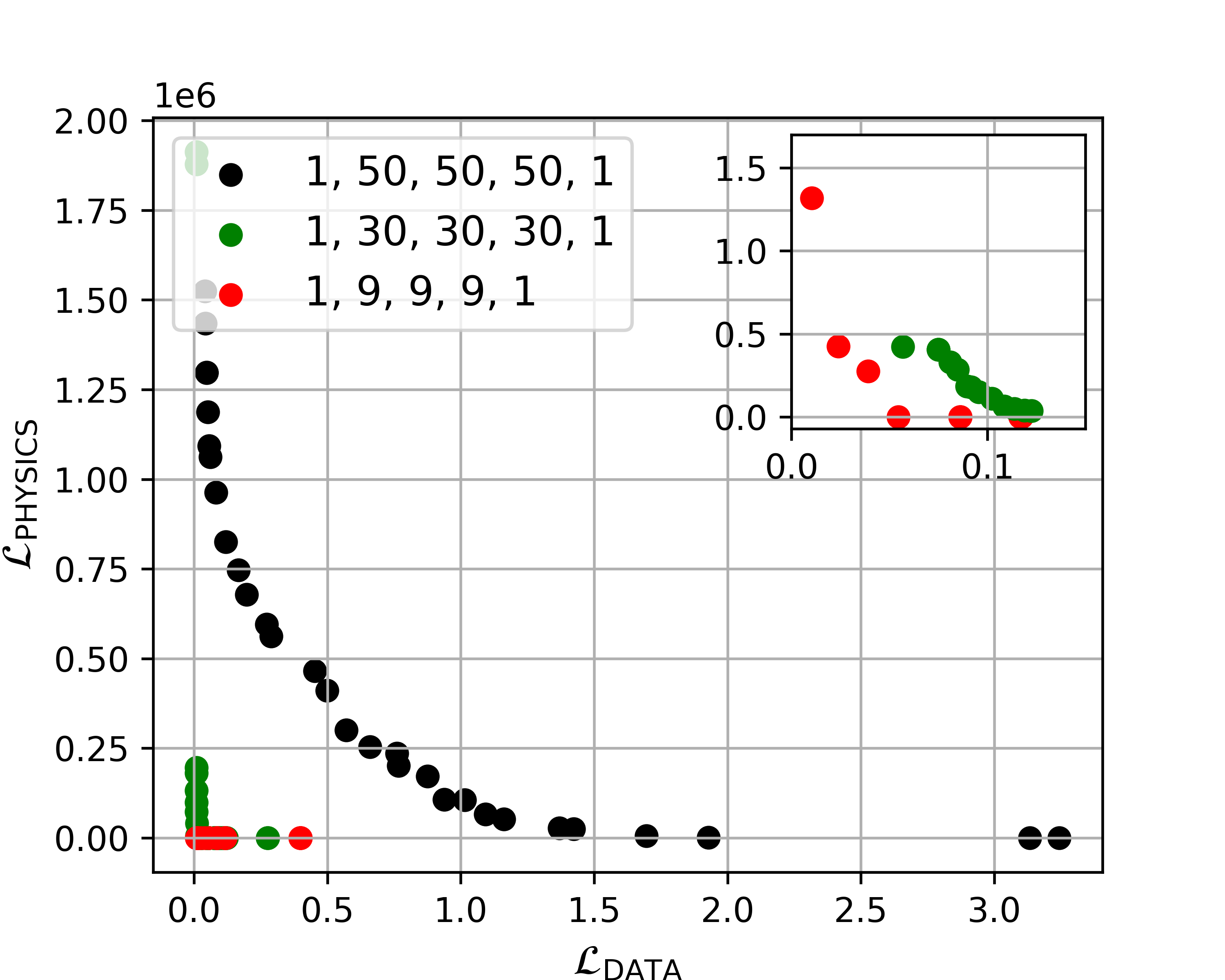}
        \caption{Different network sizes for NSGA-II.}
        \label{fig:nsga2_ntwrk_comp}
    \end{subfigure}
    \caption{Comparing the Pareto fronts for the logistic equation obtained by (a) WS, MGDA and NSGA-II, (b) increasing size of network architectures for NSGA-II.}
    \label{fig:comparisons}
\end{figure}
Even though we can compute the entire front by varying $\alpha$, the appropriate choice and spacing of the weights is extremely challenging. Through trial and error, we ended at a logarithmically scaled (with a base of 80) set for $\alpha$. Compared to the WS method, MGDA is a parameter-free MOO technique where we do not need to choose such a weight parameter (subsection \ref{subsec:MGDA}). For each random initialization in the feasible space, MGDA delivers a point on the Pareto front. On the downside, the lack of such a parameter means that we can not control where we land. Furthermore, to mitigate the effect of differences in magnitudes of gradients, it is important to normalize the gradients in each iteration before calculating the descent direction. Figure \ref{fig:mgda_fronts} shows the Pareto fronts for the logistic equation obtained from MGDA, which are convex, hence validating the empirical results obtained from the WS method. We observe that in Figure \ref{fig:mgda_norm} where gradients are normalized, the points on the front are more evenly distributed compared to Figure \ref{fig:mgda_nonorm} where the gradients have not been normalized. We also notice that apart from the non-dominated points that constitute the Pareto front, we also have dominated points. Such points are a consequence of the non-convergence of solutions occurring due to unsuitable initialization.

Figure \ref{fig:ws_Vs_mgda_Vs_nsga2} shows Pareto fronts obtained using WS, MGDA, and {NSGA-II}. While both WS and MGDA outperform {NSGA-II}, the WS method finds a large region of the front compared to MGDA. Even though EAs are global optimization methods, the algorithm appears to find a locally optimal solution only, in our case. Furthermore, we observed that compared to gradient-based methods, {NSGA-II} encountered overfitting more frequently. While we were able to define a stopping criterion (see subsection \ref{subsec:convergence}) for WS and MGDA, it is not straightforward to apply this to \mbox{NSGA-II}, as we can not keep track of individual points due to recombination. For more details about overfitting in evolutionary algorithms, refer to \cite{10.1007/978-3-642-29139-5_19, 10.1007/978-3-642-37207-0_7, inproceedings}. In Figure \ref{fig:nsga2_ntwrk_comp}, we observe that as we increase the size of the DNN, the performance of NSGA-II decreases drastically. This finding supports what other studies have also found \cite{amakor2024multiobjective, hotegni2024multiobjective}, i.e., that evolutionary algorithms are less suited to train deep neural networks due to the large number of optimization variables.
\textcolor{\revisioncolor}{In SOO, method selection is simpler, as there is only a single objective to optimize. In contrast, MOO requires careful selection of algorithms, taking into account additional factors such as the dimensionality of the objective space, the nature of the Pareto front (e.g., convexity or non-convexity), ease of implementation, and the computational cost associated with thoroughly exploring the solution space to capture the Pareto front accurately.}\\
\textcolor{\revisioncolor} {\paragraph{\textbf{Suggested Solution}} Based on the problem at hand and the strengths and weaknesses of different methods, one should choose an optimization method accordingly. A brief summary of methods together with their strengths and weaknesses is listed in Table \ref{tab:comparison_moo_methods}. A detailed overview of other existing methods is given in subsection \ref{subsec:other_methods}}
\subsection{Convergence}
\label{subsec:convergence}
\begin{figure}[htbp]
    \centering
    \begin{subfigure}{.495\textwidth}
        \includegraphics[width=\linewidth]{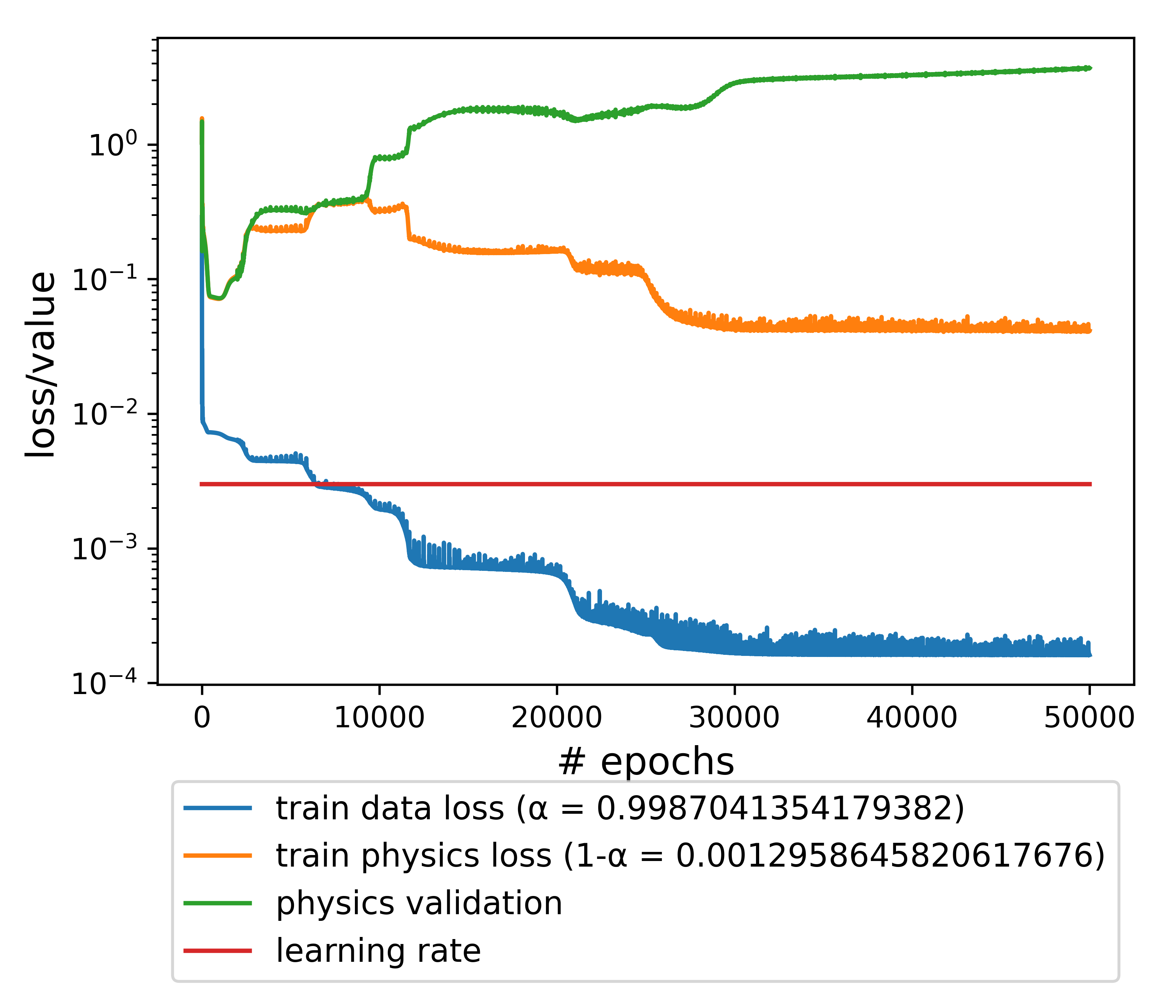}
        \caption{Low physics loss weight}
        \label{fig:loss_history_no_scheduler}
    \end{subfigure}
    \hfill
    \begin{subfigure}{.495\textwidth}
        \includegraphics[width=\linewidth]{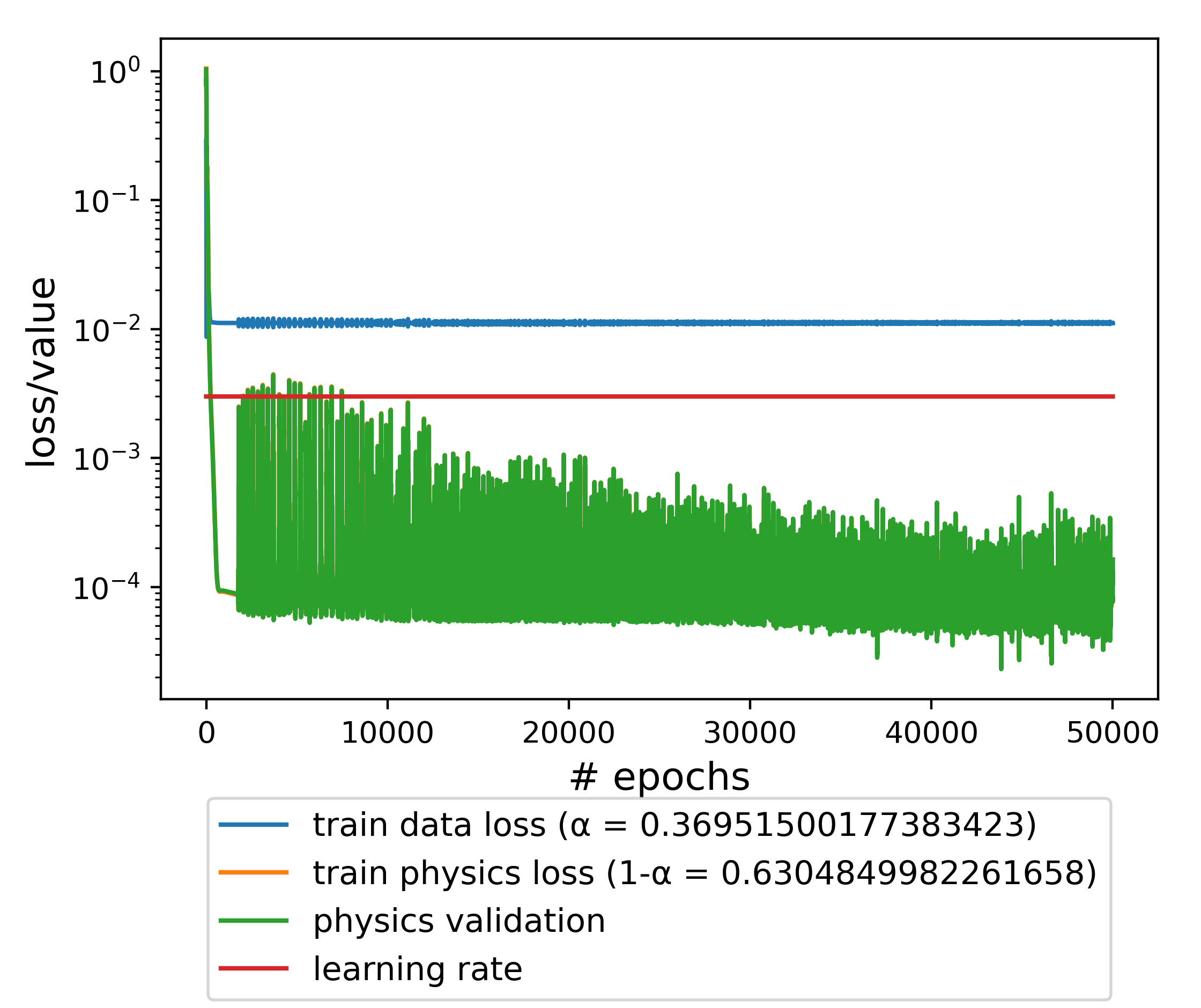}
        \caption{High physics loss weight}
        \label{fig:loss_history_no_scheduler18}
    \end{subfigure}
\caption{Loss histories over 50,000 epochs on logistic equation with validation loss tracking and constant learning rate}
\end{figure}

Neural networks require multiple training iterations, or epochs, to gradually refine their approximation of the desired function. Generally, we want to stop training when the model achieves optimal generalization, meaning that further training ceases to produce significant improvements without risking overfitting. To illustrate this concept, we consider the simpler logistic equation problem (\ref{sec:logisitc_equation}) where we calculate the Pareto front using the WS method. A validation physics loss calculated at collocation points located in between the training points (for physics and data) serves as an indicator for overfitting behavior. In Figure \ref{fig:loss_history_no_scheduler} the training loss evolution using a constant learning rate is shown. We observe that later in training, the losses start to jump unpredictably. We believe this behavior is caused by a too high learning rate for that stage of training, indicating the algorithm found a point close to a minimum. In this case, training for a fixed number of epochs can lead to stopping a model within such a spike, which would render the analysis of the results less meaningful. We also observed that this jumpy behavior was more pronounced and started earlier in training in the models with higher physics loss weight (low $\alpha$), which can be seen in the example given in Figure \ref{fig:loss_history_no_scheduler18}. This indicates that a model with high physics loss approaches the minimum faster. To mitigate the problem of the spikes in training loss, we deploy a learning rate scheduler that lowers the learning rate when the loss doesn't improve for a certain number of epochs, which leads to the smooth loss evolution seen in Figure \ref{fig:loss_history}. For the second problem of different convergence rates with different $\alpha$ we deploy the validation loss mentioned above. We stop a model when the validation loss exceeds the (physics-)training loss by more than a fixed threshold ($0.1$) for the last time in training. 
\subsubsection{Problematic Plateaus}
\begin{figure}[htbp]
    \centering
    \begin{subfigure}{.495\textwidth}
        \includegraphics[width=\linewidth]{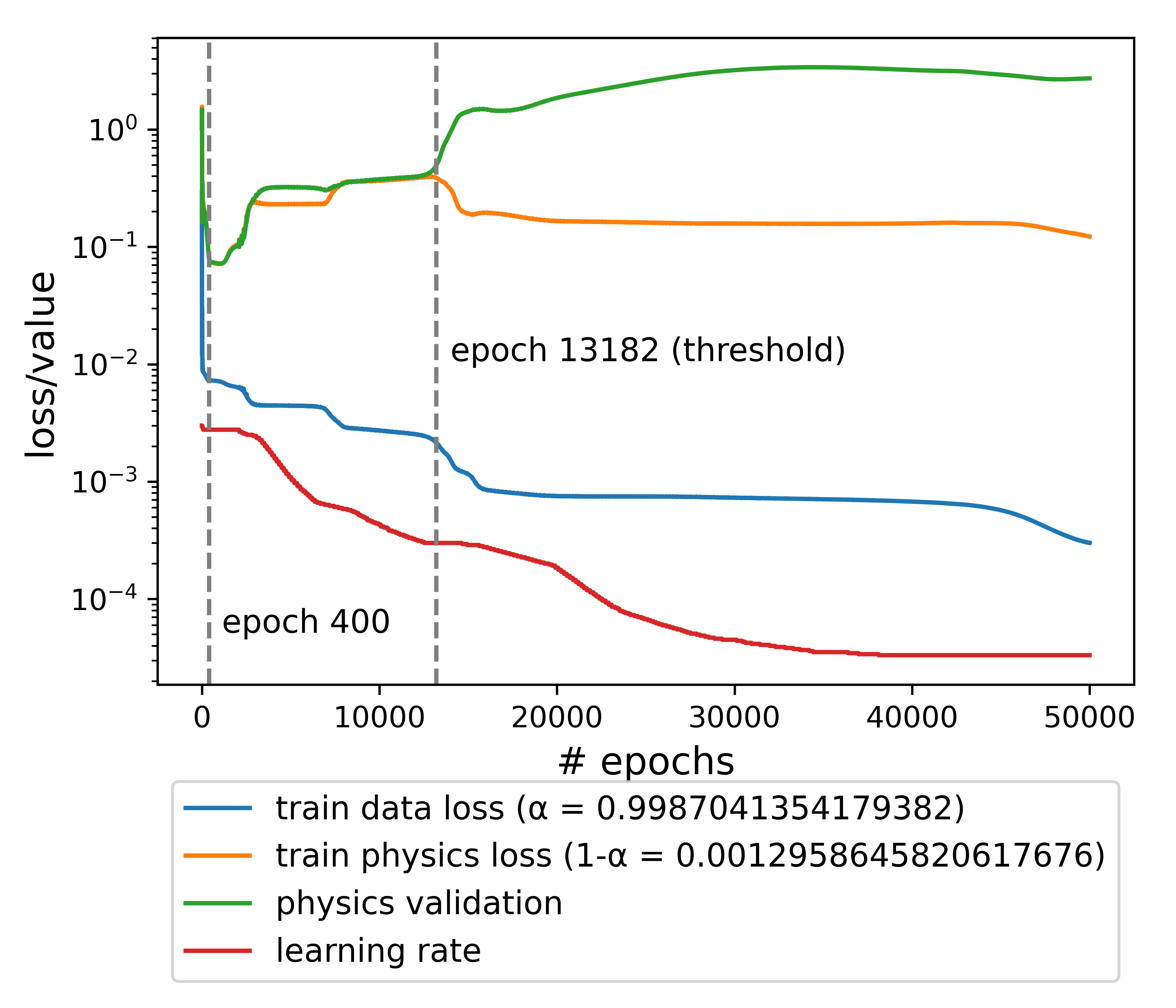}
        \caption{Loss history with learning rate scheduler}
        \label{fig:loss_history} 
    \end{subfigure}
    \hfill
    \begin{subfigure}{.495\textwidth}
        \includegraphics[width=\linewidth]{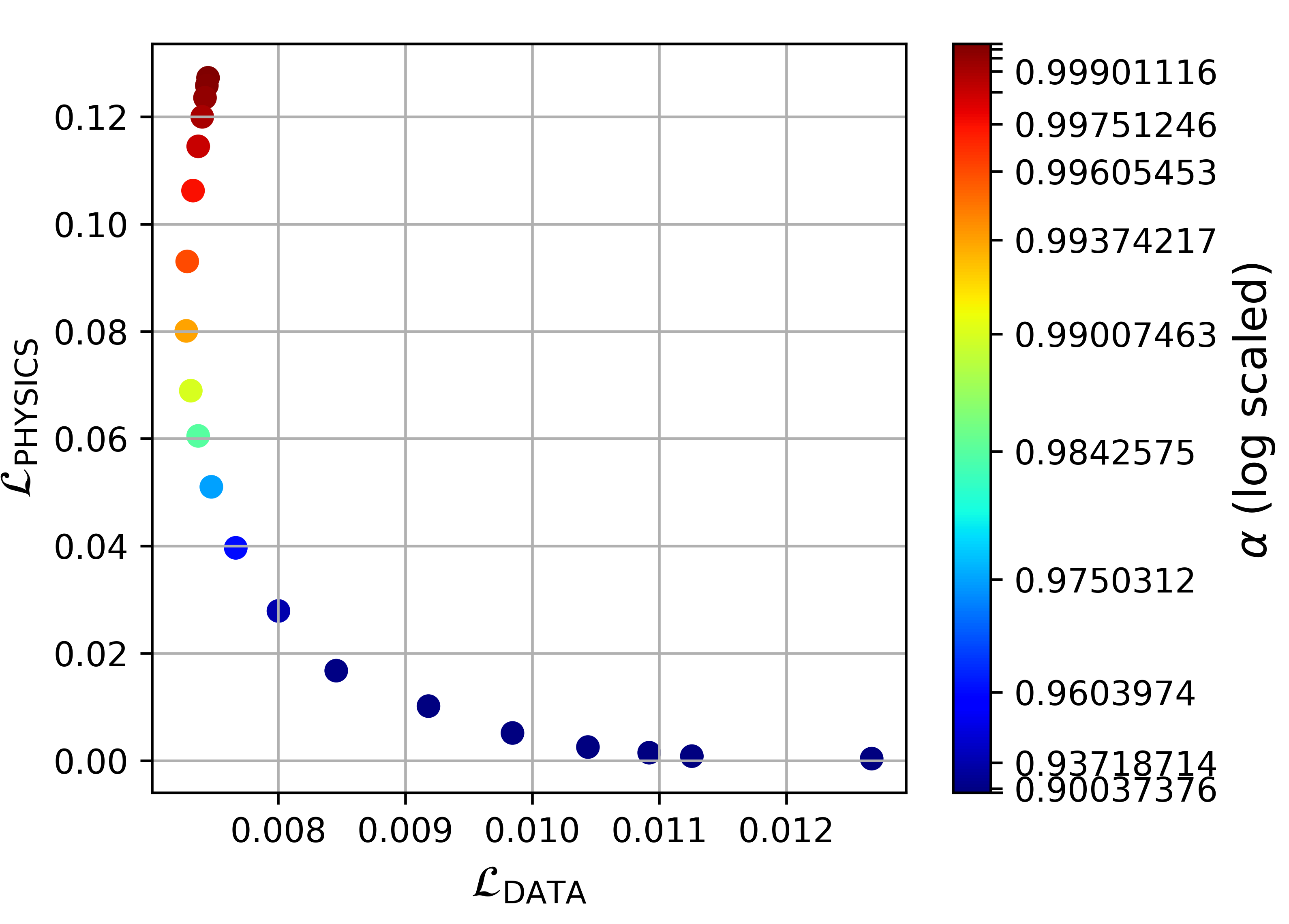}
        \caption{Solution front obtained after 400 epochs}
        \label{fig:pareto_log_400}
    \end{subfigure}
\caption{Loss history and solution front}
\end{figure}

We observe in the smooth evolution (Figure \ref{fig:loss_history}) several small plateaus in the data loss (blue) and physics loss (orange) plots. Figure \ref{fig:pareto_log_400} shows the model front after only 400 epochs of training, where the first plateau in training loss appears. Notably, several models in the upper left corner (red points) of the plot are not Pareto optimal (see Definition \ref{def:Pareto_optimal} $i)$). This means we cannot rely on a small plateau in the training evolution to indicate convergence.\\
\textcolor{\revisioncolor} {\paragraph{\textbf{Suggested Solution}} One should make sure to detect overfitting, which can be done using a threshold on a validation physics-loss or similar techniques. Also, one has to make sure to train the models to completion, while not overfitting. To deal specifically with the plateau-behavior in training, some research hints at different training algorithms that might be able to circumvent the problem \cite{ainsworth2021plateau, ainsworth2022active}.}

%% file: chapters/theory/TikzFigures/TypesOfParetoFronts.tex
 \begin{figure}[H]
         \begin{subfigure}[b]{0.24\textwidth}
             \centering
             \begin{tikzpicture}[scale=.6]
            \draw[black, thick, -stealth] (-.5  ,0.08) -- (4 ,0.08);
            \draw[black, thick, -stealth] (0,-.5 + 0.08) -- (0,4 + 0.08);
    
            \node at (3.66, -.33) {$\Loss_1$};
            \node at (-.33, 3.66) {$\Loss_2$};
            
    \draw [fill=gray!40] plot [smooth cycle, tension = .5] coordinates {
    (2, .58) (3, 1) (3.75, 2.5) (3.5, 3.5) (2.5, 3.75) (1, 3.5) (.6, 3.1) (.5, 2.5) (0.6, 2) (.75, 1.5) (1, 1.15) (1.25, .9) (1.5, .7)
    } node at (2.9,3.1)[ scale=.9] {$\mathbfcal{L}(\R^n)$};
    
    \begin{scope}
     \clip (0,0) rectangle (1.9 ,2.5);
    \draw [color = black!30!green, line width=0.4mm] plot [smooth cycle, tension = .5] coordinates {
    (2, .58) (3, 1) (3.75, 2.5) (3.5, 3.5) (2.5, 3.75) (1, 3.5) (.6, 3.1) (.5, 2.5) (0.6, 2) (.75, 1.5) (1, 1.15) (1.25, .9) (1.5, .7)
    };
    \end{scope}
    \fill[black!30!green] (1.9, .57) circle (.08);
    \fill[black!30!green] (.5, 2.5) circle (.08);
    
    \end{tikzpicture}
             \caption{}
             \label{fig:typesofFronts(a)}
        \end{subfigure}
        \hfill
        \begin{subfigure}[b]{0.24\textwidth}
            \centering
            \begin{tikzpicture}[scale=.6]
        \draw[black, thick, -stealth] (-.5,0.3) -- (4,0.3);
        \draw[black, thick, -stealth] (0,-.5 + 0.3) -- (0,4 + 0.3);

        \node at (3.66, -.33 + .3) {$\Loss_1$};
        \node at (-.33, 3.66 + .3) {$\Loss_2$};

\draw [fill=gray!40] plot [smooth cycle, tension = .5] coordinates {
(1.5, 4) (0.79, 3.71) (.6, 3) (0.79, 2.29) (1.5, 2.1) (1.9, 1.9) (2.1, 1.5) (2.29 + .1, 0.79 + .3) (3, .6 + .3) (3.71, 0.79 + .3) (4, 1.5) (4,3) (3.71, 3.71) (3, 4)
} node at (3.2,3.3)[ scale=.9] {$\mathbfcal{L}(\R^n)$};
\begin{scope}
 \clip (0,0.3) rectangle (3,3);
\draw [color = black!30!green, line width=0.4mm] plot [smooth cycle, tension = .5] coordinates {
(1.5, 4) (0.79, 3.71) (.6, 3) (0.79, 2.29) (1.5, 2.1) (1.9, 1.9) (2.1, 1.5) (2.29 + .1, 0.79 + .3) (3, .6 + .3) (3.71, 0.79 + .3) (4, 1.5) (4,3) (3.71, 3.71) (3, 4)
};
\end{scope}
\fill[black!30!green] (.6, 3) circle (.08);
\fill[black!30!green] (3, .9) circle (.08);
    
    \end{tikzpicture}
         \caption{}
         \label{fig:typesofFronts(b)}
     \end{subfigure}
     \hfill
     \begin{subfigure}[b]{0.24\textwidth}
         \centering
        \begin{tikzpicture}[scale=.6]
                    \draw[black, thick, -stealth] (-.5  ,0) -- (4 ,0);
                    \draw[black, thick, -stealth] (0,-.5) -- (0,4);
            
                    \node at (3.66, -.33) {$\Loss_1$};
                    \node at (-.33, 3.66) {$\Loss_2$};
                    
            \draw [fill=gray!40] plot [smooth cycle, tension = .5] coordinates {
            (2, .45) (3, .75) (3.55, 2) (3.5, 2.75) (3, 3.33) (2, 3.75) (1, 3.25) (.5, 2) (.75, 1.45) (1.2, 1.6) (1.5, 2.25) (1.8, 2.3) (1.9, 2) (1.8, 1.7) (1.6, 1.5) (1.5, 1.15) (1.525, 0.8)
            } node at (2.55,2.9)[ scale=.9] {$\mathbfcal{L}(\R^n)$};
            
            \begin{scope}
             \clip (0,0) rectangle (.825, 1.85);
            \draw [color = black!30!green, line width=0.4mm] plot [smooth cycle, tension = .5] coordinates {
            (2, .45) (3, .75) (3.55, 2) (3.5, 2.75) (3, 3.33) (2, 3.75) (1, 3.25) (.5, 2) (.75, 1.45) (1.2, 1.6) (1.5, 2.25) (1.8, 2.3) (1.9, 2) (1.8, 1.7) (1.6, 1.5) (1.5, 1.15) (1.525, 0.8)
            };
            \end{scope}
            \fill[black!30!green] (.5, 1.85) circle (.08);
            \fill[black!30!green] (.825, 1.425) circle (.08);

            \begin{scope}
             \clip (0,0) rectangle (2, 0.9);
            \draw [color = black!30!green, line width=0.4mm] plot [smooth cycle, tension = .5] coordinates {
            (2, .45) (3, .75) (3.55, 2) (3.5, 2.75) (3, 3.33) (2, 3.75) (1, 3.25) (.5, 2) (.75, 1.45) (1.2, 1.6) (1.5, 2.25) (1.8, 2.3) (1.9, 2) (1.8, 1.7) (1.6, 1.5) (1.5, 1.15) (1.525, 0.8)
            };
            \end{scope}
            \fill[black!30!green] (1.49, 0.9) circle (.08);
            \fill[black!30!green] (2, .45) circle (.08);
        \end{tikzpicture}
         \caption{}
         \label{fig:typesofFronts(c)}
     \end{subfigure}
     \hfill
        \begin{subfigure}[b]{0.24\textwidth}
        \centering
        \begin{tikzpicture}[scale=.6]
        \draw[black, thick, -stealth] (-.5  ,0) -- (4 ,0);
        \draw[black, thick, -stealth] (0,-.5) -- (0,4);

        \node at (3.66, -.33) {$\Loss_1$};
        \node at (-.33, 3.66) {$\Loss_2$};

        \draw [fill=gray!40] plot [smooth cycle, tension = .5] coordinates {
        (1, 1.5) (1.5, 1.6) (2,2) (2.375, 2.4) (2.75, 2.5) (2.9, 2.4) (3, 2) (3.12, 1.725) (3.4, 1.6) (3.55, 1.8) (3.65, 2.5) (3.5, 3) (3, 3.5) (2, 3.75) (1, 3) (0.55, 2.15) (0.646, 1.646)
        } node at (2.55,2.9)[ scale=.9] {$\mathbfcal{L}(\R^n)$};
        
        \draw [fill=gray!40] plot [smooth cycle, tension = .5] coordinates {
        (3, .5) (3.5, 0.75) (3.72, 2) (3.475, 3) (2.5, 3.25) (1.5, 3) (1.1, 2.5) (1.146, 2.146) (1.5, 2.05) (2, 2.5) (2.3, 2.4) (2.4, 2.2) (2.4, 1.9) (2.25, 1.1) (2.5, .6)
        } node at (2.55,2.9)[ scale=.9] {$\mathbfcal{L}(\R^n)$};
        
        \draw plot [smooth cycle, tension = .5] coordinates {
        (1, 1.5) (1.5, 1.6) (2,2) (2.375, 2.4) (2.75, 2.5) (2.9, 2.4) (3, 2) (3.12, 1.725) (3.4, 1.6) (3.55, 1.8) (3.65, 2.5) (3.5, 3) (3, 3.5) (2, 3.75) (1, 3) (0.55, 2.15) (0.646, 1.646)
        };
        
        \begin{scope}
         \clip (0,0) rectangle (1, 2.05);
        \draw [color = black!30!green, line width=0.4mm] plot [smooth cycle, tension = .5] coordinates {
        (1, 1.5) (1.5, 1.6) (2,2) (2.375, 2.4) (2.75, 2.5) (2.9, 2.4) (3, 2) (3.12, 1.725) (3.4, 1.6) (3.55, 1.8) (3.65, 2.5) (3.5, 3) (3, 3.5) (2, 3.75) (1, 3) (0.55, 2.15) (0.646, 1.646)
        };
        \end{scope}
        \fill[black!30!green] (1, 1.5) circle (.08);
        \fill[black!30!green] (0.55, 2.05)  circle (.08);
        
        \begin{scope}
         \clip (2.75, 2.5) rectangle (3.4, 1.6);
        \draw [color = blue, line width=0.4mm] plot [smooth cycle, tension = .5] coordinates {
        (1, 1.5) (1.5, 1.6) (2,2) (2.375, 2.4) (2.75, 2.5) (2.9, 2.4) (3, 2) (3.12, 1.725) (3.4, 1.6) (3.55, 1.8) (3.65, 2.5) (3.5, 3) (3, 3.5) (2, 3.75) (1, 3) (0.55, 2.15) (0.646, 1.646)
        };
        \end{scope}
        \fill[blue] (2.75, 2.5) circle (.08);
        \fill[blue]  (3.35, 1.6) circle (.08);
        
        \begin{scope}
         \clip (2.25, 1.1) rectangle (3, .4);
        \draw [color = black!30!green, line width=0.4mm] plot [smooth cycle, tension = .5] coordinates {
        (3, .5) (3.5, 0.75) (3.72, 2) (3.475, 3) (2.5, 3.25) (1.5, 3) (1.1, 2.5) (1.146, 2.146) (1.5, 2.05) (2, 2.5) (2.3, 2.4) (2.4, 2.2) (2.4, 1.9) (2.25, 1.1) (2.5, .6)
        };
        \end{scope}
        \fill[black!30!green] (2.25, 1.1) circle (.08);
        \fill[black!30!green] (3, .5)  circle (.08);
        
        \begin{scope}
         \clip (1.0, 2.5) rectangle (1.5, 2.0);
        \draw [color = blue, line width=0.4mm] plot [smooth cycle, tension = .5] coordinates {
        (3, .5) (3.5, 0.75) (3.72, 2) (3.475, 3) (2.5, 3.25) (1.5, 3) (1.1, 2.5) (1.146, 2.146) (1.5, 2.05) (2, 2.5) (2.3, 2.4) (2.4, 2.2) (2.4, 1.9) (2.25, 1.1) (2.5, .6)
        };
        \end{scope}
        \fill[blue] (1.1, 2.5) circle (.08);
        \fill[blue] (1.5, 2.05)  circle (.08);
        \end{tikzpicture}
         \caption{}
         \label{fig:typesofFronts(d)}
     \end{subfigure}
        \caption{Different types of Pareto fronts.}
        \label{fig:types_pareto_fronts}
\end{figure}

%% file: chapters_post_feedback/conclusion.tex
We have highlighted key difficulties associated with the application of MOO in DL. Our exploration began with the fundamentals of MOO, focusing mostly on the WS method and MGDA. Using both approaches, we illustrate the approximation of the Pareto fronts, navigating the weighting between the data loss and physics loss within PINNs for the modeling of differential equations. We also compared the results obtained with NSGA-II which is an evolutionary algorithm. Although we used MOO for PINNs as an illustration, the challenges and solutions we've discussed are broadly relevant to any DL scenario involving MOO. The pitfalls identified and the strategies proposed for overcoming them can be universally applied to improve the effectiveness of MOO in various DL contexts, making our insights valuable across a wide range of applications.

Our analysis in subsection \ref{subsec:identifyingfront} reveals how a basic misinterpretation of MOO can lead to erroneous identification of non-dominated points as parts of the Pareto front. \textcolor{\revisioncolor}{In subsection \ref{subsec:conflictingobjectives}, we emphasize how important conflicts in objectives are for MOO and point out possible misconceptions in previous work in that regard. Additionally, we note that the expressivity of the models can influence the trade-off between objectives.} In Subsection \ref{subsec:rightscale} we have highlighted the importance of choosing appropriate scales for visualization, showing how a convex Pareto front might misleadingly appear non-convex when plotted on a logarithmic scale, potentially leading to incorrect interpretations. In subsection \ref{subsec:knowyourmethod}, we stress the need for an in-depth understanding of the chosen MOO methodologies. We construct the Pareto fronts via the WS, MGDA, and NSGA-II, illuminating the merits and drawbacks of these methods. Moreover, we offer a strategy for selecting an MOO approach based on the specific problem at hand. Guaranteeing the optimization method's adequate convergence is vital, as discussed in subsection \ref{subsec:convergence}.

Our research not only clarifies prevalent misunderstandings in the application of MOO within DL, but also lays the groundwork for future explorations, aiming to improve the efficacy and precision of MOO in sophisticated neural network designs. These include adapting more existing MOO techniques for DL and exploring Pareto fronts for more than two objectives, which could reveal more pitfalls yet unknown to our explorations in this paper. For example, one could add sparsity as an additional objective apart from the data loss and the physics loss in PINNs.
